\documentclass[11pt]{article}

\usepackage[preprint]{acl} 

\usepackage{times}
\usepackage{latexsym}

\usepackage[T1]{fontenc}

\usepackage[utf8]{inputenc}

\usepackage{microtype}

\usepackage{inconsolata}

\usepackage{graphicx}

\usepackage{amsmath}
\usepackage{amsfonts}
\usepackage[linesnumbered,ruled,vlined]{algorithm2e}
\usepackage{multirow}
\usepackage{adjustbox}
\SetKwInput{kwInit}{Init}
\SetKwInput{KwRequire}{Require}
\usepackage{graphicx, caption, subcaption}
\usepackage{booktabs}
\usepackage{booktabs}
\usepackage{multirow}
\usepackage{colortbl} 
\usepackage{xcolor}   

\definecolor{lightblue}{rgb}{0.9, 0.95, 1.0}

\definecolor{LightBlue}{rgb}{0.88,0.95,1.0} 

\usepackage{booktabs}
\usepackage{multirow}
\usepackage{tabularx}
\usepackage{graphicx}
\usepackage{caption}
\usepackage{xspace}

\usepackage{booktabs}
\usepackage{multirow}
\usepackage{graphicx} 
\usepackage{adjustbox}       
\usepackage{colortbl}        
\usepackage{xcolor}
\usepackage{multirow}        
\usepackage{booktabs}        
\definecolor{LightBlue}{rgb}{0.8, 0.9, 1}

\usepackage{booktabs}
\usepackage{xcolor}
\usepackage{colortbl} 
\usepackage{amsmath}  
\usepackage{geometry} 
\usepackage{hyperref}
\usepackage{subcaption} 

\definecolor{failbg}{rgb}{1.0, 0.92, 0.92} 
\definecolor{passbg}{rgb}{0.92, 1.0, 0.92} 
\definecolor{highlight}{rgb}{1.0, 1.0, 0.6} 

\newtheorem{theorem}{Theorem}[section]

\newtheorem{definition}[theorem]{Definition}

\title{Do Domain-specific Experts exist in MoE-based LLMs?}

\author{
    Giang Do\thanks{Corresponding author} \quad Hung Le \quad Truyen Tran  \\
    Applied Artificial Intelligence Intiative (A2I2), Deakin University \\
    \texttt{\{truong.do,thai.le,truyen.tran\}@deakin.edu.au}\\
  }

\begin{document}
\maketitle
\begin{abstract}
In the era of Large Language Models (LLMs), the Mixture of Experts (MoE) architecture has emerged as an effective approach for training extremely large models with improved computational efficiency. This success builds upon extensive prior research aimed at enhancing expert specialization in MoE-based LLMs. However, the nature of such specializations and how they can be systematically interpreted remain open research challenges. In this work, we investigate this gap by posing a fundamental question: \textit{Do domain-specific experts exist in MoE-based LLMs?} To answer the question, we evaluate ten advanced MoE-based LLMs ranging from 3.8B to 120B parameters and provide empirical evidence for the existence of domain-specific experts. Building on this finding, we propose \textbf{Domain Steering Mixture of Experts (DSMoE)}, a training-free framework that introduces zero additional inference cost and outperforms both well-trained MoE-based LLMs and strong baselines, including Supervised Fine-Tuning (SFT). Experiments on four advanced open-source MoE-based LLMs across both target and non-target domains demonstrate that our method achieves strong performance and robust generalization without increasing inference cost or requiring additional retraining. Our implementation is publicly available at \url{https://github.com/giangdip2410/Domain-specific-Experts}.

\end{abstract}

\section{Introduction}
\label{intro}

Large Language Models (LLMs) have achieved remarkable success in Natural Language Processing (NLP) ~\citep{NEURIPS2020_1457c0d6, du_glam_2022, fedus_switch_2022}, Computer Vision~\citep{riquelme2021scalingvisionsparsemixture, shen-etal-2023-scaling}, and multimodal applications~\citep{zhan-etal-2024-anygpt,10887014}. This progress is largely driven by scaling laws~\citep{kaplan2020scalinglawsneurallanguage}, which indicate that LLM performance strongly correlates with model scale. Mixture of Experts (MoE) architectures~\citep{shazeer2017outrageously} are particularly well suited for scaling models under a fixed computational budget, making them an effective paradigm for building large-capacity models efficiently~\citep{qwen_moe,dai2024deepseekmoeultimateexpertspecialization}.

The success of MoE-based LLMs is driven by extensive work on improving expert specialization to enable efficient training and fine-tuning~\citep{krishnamurthy2023improvingexpertspecializationmixture,dai2024deepseekmoeultimateexpertspecialization,wang2024letexpertsticklast}. Yet, the nature and interpretability of such specializations remain open questions. In this work, we investigate this gap by posing a fundamental question: \textit{Do domain-specific experts exist in MoE-based LLMs?}
To tackle this problem, we conduct a systematic evaluation of recent open-source MoE-based LLMs across diverse parameter scales, ranging from large models such as GPT-OSS-120B~\citep{openai2025gptoss120bgptoss20bmodel} (120B parameters) to compact models such as Phi-tiny-MoE~\citep{li2025slimmoestructuredcompressionlarge} (3.8B parameters).

Building on these findings, we introduce \textbf{Domain Steering Mixture of Experts (DSMoE)}, a training-free framework that guides MoE-based LLMs toward improved performance while incurring no additional inference cost and requiring no further retraining. We conduct extensive experiments on four state-of-the-art MoE-based LLMs, including GPT-OSS~\citep{openai2025gptoss120bgptoss20bmodel} and Qwen3MoE~\cite{yang2025qwen3technicalreport}, across both target and non-target domains. Evaluations on challenging reasoning benchmarks such as MMLU-Pro~\citep{wang2024mmluprorobustchallengingmultitask} and GPQA-Diamond~\citep{rein2023gpqagraduatelevelgoogleproofqa} demonstrate that our method achieves strong performance while exhibiting robust generalization without increasing inference cost or requiring additional retraining. In summary, this paper makes three key contributions:

\begin{table*}[t]
\centering

\begin{tabular}{p{0.15\textwidth} p{0.80\textwidth}}
\toprule
\textbf{Question} & There exist real numbers $x$ and $y$, both greater than 1, such that $\log_x(y^x)=\log_y(x^{4y})=10$. Find $xy$. \\
\midrule

\textbf{Baseline} \newline (GPT-OSS-20B) & \cellcolor{failbg} 
\textit{Response:} We need to solve for $x, y > 1$. [...]
Equation (1): $y^x = x^{10} \implies b e^a = 10a$. 
Equation (2): $x^{4y} = y^{10} \implies 4 a e^b = 10b$.
We can attempt to solve for $a,b$. From (I): $b = 10 a / e^a$. Plug into (II): $4 a e^{10 a / e^a} = 10 b$. 
\textbf{That seems messy.} Alternatively, divide (I) by (II): $e^{a-b} = 4 a^2 / b^2$. [...]
We can attempt to solve for $t$ numerically. Let's try $t=1$, $t=2$, $t=0.5$... [loops through 20+ numerical guesses]. 
So left side negative until $t \approx 2.5$, right side positive. \textbf{So no crossing? So equality cannot hold.} 
\newline
\textbf{Predicted Answer:} 0 \quad \textcolor{red}{\textbf{(Incorrect)}} \\

\midrule

\textbf{DSMoE} \newline (Ours) & \cellcolor{passbg} 
\textit{Response:} We have $x \ln y = 10 \ln x$ and $4y \ln x = 10 \ln y$.
Let $u = \ln x, v = \ln y$. Equations: $e^u v = 10 u$ and $4 e^v u = 10 v$.
Dividing equations leads to $u - v = \ln 4 + 2 \ln u - 2 \ln v$.
[...]
Let's attempt to find integer solution for $xy$. Suppose $xy=25$. 
We have $u+v = \ln 25 \approx 3.218$.
Solving numerically for $u$ with sum $s=\ln 25$: $e^u (s-u) = 10u$.
We find $u \approx 0.4275$. Checking second equation: $4 e^{s-u} u = 10(s-u)$.
Left side: $27.904$. Right side: $27.913$. \textbf{Very close. So indeed $u$ solves both. So $xy=25$ is consistent.}
\newline
\textbf{Predicted Answer:} 25 \quad \textcolor{green!60!black}{\textbf{(Correct)}} \\

\bottomrule
\end{tabular}
\caption{Qualitative comparison on a challenging math problem - \textbf{AIME}~\citep{maa_aime} 2024. The baseline (GPT-OSS-20B) fails to resolve the system of equations, getting stuck in numerical approximation. In contrast, DSMoE successfully identifies the integer relationship $xy=25$ through effective hypothesis testing.}
\label{tab:qualitative_math}
\end{table*}

\begin{itemize}
    \item We formulate the interpretability of MoE-based LLMs as the problem of verifying domain-specific experts.
    \item We empirically evaluate ten advanced MoE-based LLMs across a wide range of model scales, providing evidence for the existence of domain-specific specialization.
    \item We propose \textbf{Domain Steering MoEs (DSMoE)}, a training-free framework that achieves strong performance and generalization without incurring additional inference or retraining costs.
\end{itemize}


\section{Related work}
\label{related}

{\bf Expert Specialization. } Mixture of Experts (MoE) models~\citep{jacobs1991,jordan1994} have gained significant traction in large language models and have since been widely applied across domains such as natural language processing, computer vision, and speech recognition~\citep{jiang2024mixtralexperts,NEURIPS2022_2f00ecd7,NEURIPS2021_48237d9f}. However, ensuring that experts acquire non-overlapping and specialized knowledge remains challenging~\citep{dai2024deepseekmoeultimateexpertspecialization,chi2022representation}. To address this challenge, prior work has followed two main research directions: 
(1) modifying the MoE architecture and 
(2) improving the routing mechanism. 
Following the first direction, DeepSeekMoE~\citep{dai2024deepseekmoeultimateexpertspecialization} promotes expert specialization through fine-grained expert segmentation and the introduction of shared experts, while $\mu\mathrm{MoE}$~\citep{NEURIPS2024_5eeb693f} achieves specialization by performing implicit computation over prohibitively large weight tensors entirely in a factorized form. 
Along the second direction, various routing-based solutions have been proposed, including XMoE, which employs low-dimensional routing scores~\citep{chi2022representation}, and SMoE-dropout, which gradually activates a larger number of experts during training~\citep{chen2023sparse}. 
Other approaches, such as  StableMoE~\citep{dai2022stablemoe} and HyperRouter~\citep{do2023hyperrouter}, focus on improving router stability and robustness. Beyond architectural and routing modifications, some studies propose auxiliary loss functions to further enhance expert specialization in MoE-based LLMs~\citep{do2024simsmoesolvingrepresentationalcollapse,guo2025advancingexpertspecializationbetter}.

\noindent
\textbf{Explainable MoE.} The demand for reliable and transparent model explanations is critical across many machine learning applications, for which Mixture of Experts (MoE) architectures are particularly well suited. Interpretable MoE (IME)~\citep{ismail2023interpretablemixtureexperts} integrates MoE with deep neural networks (DNNs) to achieve both strong predictive performance and enhanced interpretability. Following a similar line of research, SMoE-VAE~\citep{nikolic2025exploringexpertspecializationunsupervised} combines MoE with variational autoencoders to improve expert-level interpretability in an unsupervised setting. More recently, MoE Lens~\citep{chaudhari2025moe} analyzes domain-specific routing patterns and demonstrates that MoE models predominantly rely on a small subset of specialized experts, with the top-weighted expert's output closely approximating the full ensemble prediction.

\noindent
\textbf{Steering LLMs.} 
Controlling the behavior of large language models (LLMs) through direct intervention on internal activations has emerged as a promising research direction. 
Prior work has proposed various activation-steering methods that modify model behavior by injecting steering signals into intermediate representations~\citep{turner2024steeringlanguagemodelsactivation,rimsky-etal-2024-steering}. 
Most existing approaches apply a steering vector to the model activations at specific layers and token positions during inference; however, such methods typically rely on locally learned steering vectors, which limits their ability to generalize across multiple domains~\citep{wang2025two}. RICE~\citep{wang2025two} recently introduces a steering method that focuses on \emph{thinking experts}. However, this approach is currently limited to extremely large reasoning models, such as DeepSeek-R1~\citep{deepseekai2025deepseekr1incentivizingreasoningcapability} and Qwen3-235B~\citep{qwen3technicalreport}. Moreover, RICE targets thinking experts that exhibit substantial variability across samples or domains, which can hinder the method's generalization capability. In contrast, our work addresses a fundamental question: \textit{Do domain-specific experts exist in MoE-based LLMs?} To this end, we conduct a systematic evaluation of recent open-source MoE-based LLMs across a wide range of parameter scales, such as GPT-OSS-120B~\citep{openai2025gptoss120bgptoss20bmodel} or Phi-tiny-MoE~\citep{li2025slimmoestructuredcompressionlarge}. Based on these conclusions, we propose \textbf{Domain Steering Mixture of Experts (DSMoE)}, a training-free framework that introduces zero additional inference cost and consistently outperforms well-trained MoE-based LLMs as well as strong baselines, including supervised fine-tuning (SFT).

\section{Methodology}
\label{method}

\subsection{Preliminaries}

\textbf{MoE-based LLM Layer.}\;
An MoE-based LLM layer replaces the standard Feedforward Network (FFN) with a Mixture of Experts module consisting of $N$ experts. Given an input $\boldsymbol{x} \in \mathbb{R}^{d}$, the layer computes a weighted aggregation of $k$ active experts:
\begin{equation}
    f^{\mathrm{MoE}}(\boldsymbol{x}) = \sum_{i \in \mathcal{K}} s_i(\boldsymbol{x}) \, \mathrm{FFN}_i(\boldsymbol{x}),
\end{equation}
where $\boldsymbol{W}_e \in \mathbb{R}^{N \times d}$ is the router embedding and $\mathcal{K} \subset \{1, \dots, N\}$ is the set of $k$ selected indices. The gating weights $\boldsymbol{s}(\boldsymbol{x})$ are derived from the router logits $\boldsymbol{l} = \boldsymbol{W}_e \boldsymbol{x}$. Depending on the architecture, sparsity is enforced either \textit{after} softmax (standard) or \textit{before} softmax (masked). Letting $\sigma(\cdot)$ denote the softmax function:
\begin{equation}
    \boldsymbol{s}(\boldsymbol{x}) = 
    \begin{cases}
       \operatorname{TopK}(\sigma(\boldsymbol{l}), k) & \text{\footnotesize(Post-Softmax)} \\
       \sigma(\operatorname{TopK}_{\text{mask}}(\boldsymbol{l}, k)) & \text{\footnotesize(Pre-Softmax)}
    \end{cases}
\label{eq:moe_routing_variants}
\end{equation}
where $\operatorname{TopK}_{\text{mask}}$ sets the logits of non-selected experts to $-\infty$ prior to normalization. Each expert $\mathrm{FFN}_i$ is a multi-layer perceptron.


\subsection{Domain-specific Token}



\begin{definition}[Common Token]
\label{def:common_token}
Let $\mathcal{D}$ denote a specific domain and $S = (s_1, s_2, \dots, s_T)$ be a sequence of tokens such that $S \in \mathcal{D}$. Let $\mathcal{M}$ be a pre-trained Large Language Model and $\mathcal{T}(\cdot, \mathcal{M})$ be the task evaluation metric.

A token $s_i$ is defined as a \textit{Common Token} if its removal results in a negligible change in the task metric, bounded by a small threshold $\epsilon$:
\begin{equation}
    \left| \mathcal{T}(S, \mathcal{M}) - \mathcal{T}(S_{\setminus \{s_i\}}, \mathcal{M}) \right| < \epsilon
\end{equation}
where $S_{\setminus \{s_i\}}$ denotes the sequence $S$ excluding token $s_i$, and $\epsilon \approx 0$ is a small scalar.
\end{definition}

\begin{definition}[Domain-specific Token]
\label{def:spec_token}
Following the notation in Definition \ref{def:common_token}, a token $s_i$ is defined as a \textit{Domain-specific Token} if its removal causes a significant degradation in the task metric, exceeding the threshold $\epsilon$:
\begin{equation}
    \left| \mathcal{T}(S, \mathcal{M}) - \mathcal{T}(S_{\setminus \{s_i\}}, \mathcal{M}) \right| \geq \epsilon
\end{equation}
This inequality implies that $s_i$ carries information crucial to the performance of model $\mathcal{M}$ on domain $\mathcal{D}$.
\end{definition}


\noindent \textit{Remark:} Unless otherwise specified, $\mathcal{T}(\cdot, \mathcal{M})$ denotes the cross-entropy loss, as this is the standard objective function for next-token prediction tasks in Large Language Models.


A direct approach to identifying domain-specific tokens satisfying Definition \ref{def:spec_token} is Leave-One-Out (LOO) cross-validation~\citep{1716307}. However, this method is computationally prohibitive for long sequences, as it scales with complexity $\mathcal{O}(N^2)$. To overcome this limitation, we leverage gradient-based attribution methods to approximate token importance~\citep{shrikumar2019learningimportantfeaturespropagating, ancona2018betterunderstandinggradientbasedattribution}. 

Specifically, let $\mathbf{e}_i \in \mathbb{R}^d$ denote the input embedding vector for token $s_i$, and let $\mathcal{L}$ be the task loss. We calculate the ranking score $r_i$ for each token as the $L_2$ norm of the element-wise product between the embedding and its gradient:

\begin{equation}
    r_i = \left\| \mathbf{e}_i \odot \nabla_{\mathbf{e}_i} \mathcal{L} \right\|_2
\end{equation}

\noindent where $\odot$ denotes the Hadamard product and $\nabla_{\mathbf{e}_i} \mathcal{L}$ is the gradient of the loss with respect to the embedding vector. High values of $r_i$ indicate tokens that significantly influence the model's output.



\begin{definition}[Domain-specific Threshold]
\label{def:d_thresh}
Let $r = (r_1, r_2, \dots, r_T)$ be the vector of token importance scores for sequence $S$. For a chosen \textit{domain-specific level} $p \in [0, 1]$ (representing the proportion of tokens considered domain-specific), we define the threshold $t_p$ as the value satisfying the empirical quantile condition:

\begin{equation}
    \frac{1}{T} \sum_{i=1}^{T} \mathbb{I}(|r_i| \le t_p) = 1 - p
\end{equation}

\noindent where $\mathbb{I}(\cdot)$ is the indicator function. The \textit{token classification function} $\mathcal{C}_{t_p}: \mathbb{R}^T \to \{\text{Common, Specific}\}^T$ is defined as:

\begin{equation}
    \mathcal{C}_{t_p}(r)_i = 
    \begin{cases} 
      \text{Common Token} & \text{if } |r_i| \le t_p \\
      \text{Domain-specific Token} & \text{otherwise}
   \end{cases}
\end{equation}
\end{definition}


\noindent \textit{Remark:} The hyperparameter $p$ is selected empirically. We find that values in the range $(0.15, 0.5)$ are effective, a setting consistent with the 15\% masking ratio used in the BERT pre-training objective~\citep{devlin-etal-2019-bert}.






\subsection{Domain-specific Expert}

We hypothesize the existence of \textit{domain-specific experts} within the Mixture-of-Experts (MoE) architecture. These experts are characterized by a dual property: they are frequently activated within a specific domain, and when activated, they exhibit a strong preference for processing domain-specific tokens rather than common tokens.






\begin{definition}[Domain-specific Expert]
\label{def:ds_expert}
Let $S$ be a sequence of tokens from domain $\mathcal{D}$, partitioned into a set of domain-specific tokens $\mathcal{S}$ and common tokens $\mathcal{C}$ (as per Definition \ref{def:d_thresh}). Let $\mathcal{E} = \{e_1, \dots, e_N\}$ be the set of $N$ experts in the model.

For each expert $e_j$, we calculate the \textit{domain-specific score} $g(e_j)$ as:
\begin{equation}
    g(e_j) = P(e_j | \mathcal{D}) \cdot \left[ P(s \in \mathcal{S} | e_j) - P(s \in \mathcal{C} | e_j) \right]
\label{eq:ranks}
\end{equation}
where $P(e_j | \mathcal{D})$ is the activation frequency of $e_j$ on domain $\mathcal{D}$, and the conditional probabilities represent the expert's token preference.

Let $K$ be a hyperparameter denoting the target number of domain-specific experts (i.e., top-$K$). We define the selection threshold $\gamma_K$ such that:
\begin{equation}
    \sum_{j=1}^{N} \mathbb{I}(g(e_j) \geq \gamma_K) = K
\end{equation}
Accordingly, the set of \textit{Domain-specific Experts} $\mathcal{E}^* \subset \mathcal{E}$ is defined as the subset of experts satisfying this condition:
\begin{equation}
    \mathcal{E}^* = \{ e_j \in \mathcal{E} \mid g(e_j) \geq \gamma_K \}
\end{equation}
\end{definition}

\noindent \textit{Remark:} The score $g(e_j)$ effectively balances two factors: the expert's global relevance to the domain (represented by $P(e_j | \mathcal{D})$) and its specialization level (represented by the difference in conditional probabilities). This penalizes experts that are active frequently but only process common, non-informative tokens.


\subsection{Domain Steering Mixture of Experts (DSMoE)}

Building upon the identification of domain-specific experts in Definition \ref{def:ds_expert}, we propose \textbf{Domain Steering Mixture of Experts (DSMoE)}. DSMoE is a training-free inference framework designed to adapt generic MoE-based Large Language Models to specific target domains by dynamically modulating the router's behavior.

Formally, let $\mathcal{E}^*$ denote the set of identified domain-specific experts. For a given input token, let $w_j$ represent the original scalar weight (or logit) assigned to expert $e_j$ by the router. We introduce a \textit{steering coefficient} $\alpha \in (0, 100)$ to amplify the contribution of domain-specific experts.

The steered routing weights $\tilde{w}_j$ are computed as follows:
\begin{equation}
    \tilde{w}_j = 
    \begin{cases} 
      \alpha \cdot w_j & \text{if } e_j \in \mathcal{E}^* \\
      w_j & \text{otherwise}
   \end{cases}
\end{equation}

\noindent After applying this steering transformation, the weights are typically re-normalized (e.g., via a Softmax function) to ensure a valid probability distribution for expert selection. This mechanism effectively biases the model's computation path towards experts specialized for the target domain without requiring parameter updates.





\section{Experiments}



We design our experiments to investigate the following four research questions: \textbf{RQ1 (Existence)} asks if domain-specific experts exist in MoE-based LLMs; \textbf{RQ2 (Effectiveness)} evaluates how effective the proposed DSMoE framework is on target domains; \textbf{RQ3 (Generalization)} examines if DSMoE maintains robust performance on non-target domains; and \textbf{RQ4 (Efficiency)} analyzes the computational cost of DSMoE compared to baselines, particularly regarding inference overhead.

\subsection{Experimental Settings}



\noindent \textbf{MoE-based LLMs.}\; To validate our hypothesis, we evaluate our method across a diverse set of state-of-the-art Mixture-of-Experts (MoE) Large Language Models, ranging in scale from 3.8B to 120B parameters. Specifically, our experimental suite includes PhiMoE-Tiny~\citep{li2025slimmoestructuredcompressionlarge}, OLMoE~\citep{muennighoff2024olmoeopenmixtureofexpertslanguage}, Qwen1.5-MoE~\citep{qwen_moe}, and DeepSeek-MoE~\citep{dai2024deepseekmoeultimateexpertspecialization}. We also include recent advanced models such as GPT-OSS-20B and GPT-OSS-120B~\citep{openai2025gptoss120bgptoss20bmodel}, ERNIE-4.5~\citep{ernie2025technicalreport}, and the Qwen3-MoE series (Instruct, Think, and Next variants)~\citep{qwen3technicalreport}. Detailed architectural specifications for all utilized models are provided in Table~\ref{tab:model_specs}.

\noindent \textbf{Baselines.}\; Since our proposed DSMoE is a training-free framework compatible with any off-the-shelf MoE-based LLMs, our main baseline is the Original MoE-based models. We also compare our approach against RICE~\citep{wang2025expertsneedsteeringthinking}, a recent state-of-the-art method for steering MoE-based reasoning models. To provide a comprehensive evaluation, we further benchmark against Supervised Fine-Tuning (SFT) using LoRA~\citep{hu2021loralowrankadaptationlarge}, which serves as the standard paradigm for domain adaptation. Throughout this paper, \emph{SFT} refers to fine-tuning using LoRA, with trainable parameters comprising 2.7\% of the total model parameters, unless stated otherwise.


\noindent \textbf{Benchmarks.}\; To demonstrate the efficacy of DSMoE, we evaluate our method on three challenging benchmarks designed to assess deep domain understanding and complex reasoning. We first utilize \textbf{MMLU-Pro}~\citep{wang2024mmluprorobustchallengingmultitask}, a robust benchmark that introduces reasoning-intensive questions with a ten-option choice set to minimize random guessing. Furthermore, we test on \textbf{GPQA Diamond}~\citep{rein2023gpqagraduatelevelgoogleproofqa}, a graduate-level dataset of such extreme difficulty that domain experts (PhDs) achieve only $\approx 65\%$ accuracy. Finally, we evaluate advanced mathematical proficiency using \textbf{AIME}~\citep{maa_aime}, a collection of problems from the American Invitational Mathematics Examination known for requiring multi-step reasoning.



\subsection{Domain-specific Experts Testing}

To verify that the domain-specific experts defined in Section~\ref{def:ds_expert} are genuinely specialized, we conduct a performance evaluation on the mathematics domain using ten advanced MoE-based LLMs. For each model, we sample 10\% of mathematics questions from the MMLU-Pro dataset. The evaluation follows the procedure described in Section~\ref{method}, consisting of three steps: (1) identifying domain-specific tokens as defined in Definition~\ref{def:spec_token}; (2) computing expert ranking scores according to Equation~\ref{eq:ranks}; and (3) applying DSMoE with $K=1$ by activating the expert with the highest ranking score.

\begin{figure*}[!t]
    \centering
    \includegraphics[width=\textwidth]{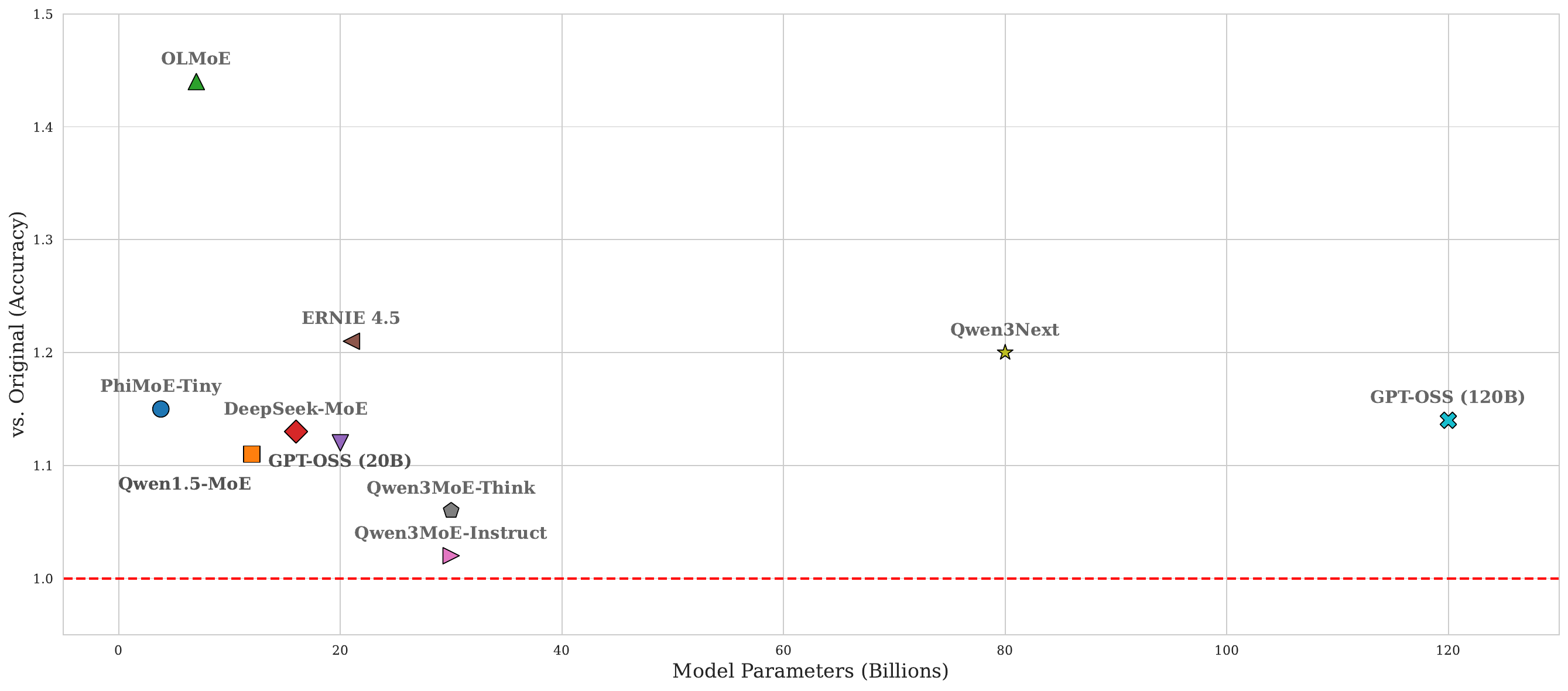}
    \caption{Evaluation of Domain-specific Experts across ten state-of-the-art MoE LLMs on the MMLU-Pro Math domain. Experiments were conducted with $K=1$ (number of specific experts) and $\alpha=3.0$ (steering coefficient). All models exhibit performance improvements compared to their original baselines, providing empirical evidence for the existence of domain-specific experts within MoE architectures. Best viewed in color.}
    \label{fig:perform_test}
\end{figure*}

For Step (1), to ensure a fair evaluation, we identify domain-specific tokens using only the question tokens, excluding answer tokens. We visualize the evaluation results by comparing DSMoE predictions with the ground-truth labels of the MMLU-Pro mathematics benchmark, as shown in Figure~\ref{fig:perform_test}. The results demonstrate that at least one domain-specific expert exists for the mathematics domain in all ten evaluated MoE-based LLMs. These findings support our hypothesis that MoE-based LLMs inherently contain domain-specific experts. Moreover, an interesting observation is that steering a single expert in MoE-based LLMs can significantly improve performance over the base model, with gains ranging from \textbf{3\%} to \textbf{45\%}.

\subsection{Target Domains Evaluation}

Conventional steering methods, such as RICE~\citep{wang2025two}, require rerunning the steering procedure separately for each domain and each dataset. In contrast, DSMoE is built upon domain-specific representations and requires identifying domain-specific tokens (Definition~\ref{def:spec_token}) and domain-specific experts (Definition~\ref{def:ds_expert}) only once. Since the MMLU-Pro dataset supports 14 domains, we adopt it as the target dataset for identifying domain-specific tokens and experts. In this work, we focus on four target domains (Math, Biology, Physics, and Chemistry); however, DSMoE can be straightforwardly extended to other domains supported by MMLU-Pro.

For each target domain, we use only question texts (without labels) from MMLU-Pro to identify domain-specific tokens and domain-specific experts. Empirically, we find that $p \in (0.15, 0.5)$ and $K$ set to approximately 1\% of the total number of experts yield strong performance. Notably, smaller values of $p$ impose a stricter criterion for domain specificity, resulting in fewer selected experts but higher confidence in their domain specialization. Table~\ref{tab:main_results} presents the comparative results across four target domains on the MMLU-Pro benchmark. Overall, DSMoE consistently outperforms both the original MoE baselines and the advanced steering method, RICE, across all evaluated models. Specifically, DSMoE yields average absolute improvements of \textbf{+1.5}, \textbf{+14.5}, \textbf{+3.6}, and \textbf{+3.7} percentage points for Qwen3-30B-Instruct, GPT-OSS-120B, Qwen3-30B-Thinking, and GPT-OSS-20B, respectively.

Our analysis highlights significant limitations in baseline methods. The results indicate that RICE, which relies heavily on thinking tokens, struggles to generalize to standard (non-reasoning) models. Furthermore, DSMoE frequently outperforms Supervised Fine-Tuning (SFT). While SFT requires updating approximately 2.7\% of the model's parameters and is inherently data-intensive, DSMoE achieves superior performance without any parameter updates. Notably, our method achieves substantial gains in challenging domains such as Mathematics and Chemistry, recording improvements of up to \textbf{+29.1} points over the original model. These findings demonstrate that selectively activating domain-specific experts is a highly efficient strategy for enhancing model performance without the computational and data costs of fine-tuning.


\begin{table}[t]
\centering
\resizebox{\columnwidth}{!}{
\begin{tabular}{lccccc}
\toprule
\textbf{Domain} & \textbf{Orig.} & \textbf{RICE} & \textbf{SFT} & \textbf{DSMoE} & \textbf{+/-} \\
\midrule
\multicolumn{6}{l}{\textit{\textbf{Qwen3-30B-Instruct}}} \\
\midrule
Math & 85.0 & 85.9 & 85.4 & \textbf{86.3} & +1.3 \\
Biology & 87.6 & 87.0 & 87.9 & \textbf{88.6} & +1.0 \\
Physics & 78.1 & 79.1 & 78.8 & \textbf{80.0} & +1.8 \\
Chemistry & 76.2 & 77.7 & 77.1 & \textbf{78.3} & +2.0 \\
\rowcolor{lightblue}
Average & 81.7 & 82.4 & 82.3 & \textbf{83.3} & +1.5 \\
\midrule
\multicolumn{6}{l}{\textit{\textbf{GPT-OSS-120B}}} \\
\midrule
Math & 74.0 & 82.8 & 77.4 & \textbf{87.5} & +13.5 \\
Biology & 72.8 & 85.8 & 86.8 & \textbf{88.1} & +15.3 \\
Physics & 82.4 & 81.8 & 80.9 & \textbf{82.6} & +0.2 \\
Chemistry & 50.2 & 78.0 & \textbf{79.8} & 79.2 & +29.1 \\
\rowcolor{lightblue}
Average & 69.8 & 82.1 & 81.2 & \textbf{84.4} & +14.5 \\
\midrule
\multicolumn{6}{l}{\textit{\textbf{Qwen3-30B-Thinking}}} \\
\midrule
Math & 75.3 & 74.5 & 72.8 & \textbf{78.3} & +3.1 \\
Biology & 73.9 & 74.3 & 73.2 & \textbf{75.2} & +1.3 \\
Physics & 58.7 & 59.4 & 61.8 & \textbf{65.8} & +7.1 \\
Chemistry & 56.5 & \textbf{59.3} & 58.7 & \textbf{59.3} & +2.8 \\
\rowcolor{lightblue}
Average & 66.1 & 66.9 & 66.6 & \textbf{69.6} & +3.6 \\
\midrule
\multicolumn{6}{l}{\textit{\textbf{GPT-OSS-20B}}} \\
\midrule
Math & 66.5 & 64.7 & 62.6 & \textbf{74.5} & +8.1 \\
Biology & 76.4 & 73.1 & 73.9 & \textbf{78.7} & +2.2 \\
Physics & 62.0 & 57.3 & 58.2 & \textbf{65.7} & +3.8 \\
Chemistry & 48.9 & 48.9 & 47.3 & \textbf{49.6} & +0.7 \\
\rowcolor{lightblue}
Average & 63.4 & 61.0 & 60.5 & \textbf{67.1} & +3.7 \\
\bottomrule
\end{tabular}%
}
\vspace{5pt}
\caption{Accuracy (\%) on the MMLU-Pro benchmark across different domains. Best results per row are highlighted in \textbf{bold}. The $\Delta$ column indicates the absolute improvement (in percentage points) of DSMoE over the original MoE-based LLMs.}
\label{tab:main_results}
\end{table}

\subsection{Non-target Domains Evaluation}

\noindent \textbf{Generalizability Evaluation.}; To demonstrate the generalizability of DSMoE, we evaluate its performance on two independent benchmarks using domain-specific experts identified via the MMLU-Pro dataset. First, Table~\ref{tab:gpqa_results} reports results on the GPQA Diamond benchmark across Biology, Physics, and Chemistry. DSMoE consistently outperforms the original MoE baselines across all evaluated models, yielding average gains ranging from \textbf{+4.8} to \textbf{+27.1} percentage points.
To further assess performance on extremely challenging reasoning tasks, we evaluate DSMoE on the American Invitational Mathematics Examination (AIME)~\citep{maa_aime}. As shown in Table~\ref{tab:aime_results}, DSMoE surpasses all baselines on both the 2024 and 2025 datasets using Qwen3-30B-Instruct and GPT-OSS-20B. Remarkably, DSMoE achieves improvements ranging from \textbf{+12.3} to \textbf{+27.3} percentage points over the baseline. These results confirm that DSMoE effectively generalizes across datasets of varying difficulty levels within a specific domain, maintaining superior performance even on competition-grade problems.

Notably, DSMoE exhibits robust transfer capabilities without systematic degradation on these unseen tasks. In many instances, it surpasses both RICE (a training-free steering baseline) and SFT (a fine-tuning method). The improvements are particularly pronounced for smaller architectures, such as GPT-OSS-20B, suggesting that selectively activating domain-relevant experts enhances fundamental scientific reasoning beyond the specific dataset used for identification. These results indicate that DSMoE effectively preserves, and often significantly enhances, cross-domain generalization while applying targeted expert steering.

\begin{table}[t]
\centering
\resizebox{\columnwidth}{!}{
\begin{tabular}{lccccc}
\toprule
\textbf{Domain} & \textbf{Orig.} & \textbf{RICE} & \textbf{SFT} & \textbf{DSMoE} & \textbf{+/-} \\
\midrule
\multicolumn{6}{l}{\textit{\textbf{Qwen3-30B-Instruct}}} \\
\midrule
Biology & 68.4 & 47.4 & 63.2 & \textbf{73.7} & +5.3 \\
Physics & 70.9 & 70.9 & 74.4 & \textbf{76.7} & +5.8 \\
Chemistry & 53.8 & 45.2 & 41.9 & \textbf{58.1} & +4.3 \\
\rowcolor{lightblue}
Average & 64.4 & 54.5 & 59.8 & \textbf{69.5} & +5.1 \\
\midrule
\multicolumn{6}{l}{\textit{\textbf{GPT-OSS-120B}}} \\
\midrule
Biology & 58.0 & 63.2 & \textbf{73.7} & 68.4 & +10.4 \\
Physics & 88.4 & 86.1 & 82.6 & \textbf{89.5} & +1.2 \\
Chemistry & 50.5 & 43.0 & 40.9 & \textbf{59.1} & +8.6 \\
\rowcolor{lightblue}
Average & 65.6 & 64.1 & 65.7 & \textbf{72.4} & +6.7 \\
\midrule
\multicolumn{6}{l}{\textit{\textbf{Qwen3-30B-Thinking}}} \\
\midrule
Biology & 52.6 & 42.1 & 21.1 & \textbf{57.9} & +5.3 \\
Physics & 55.8 & 46.5 & 54.7 & \textbf{60.5} & +4.7 \\
Chemistry & 47.3 & 20.4 & 26.9 & \textbf{51.6} & +4.3 \\
\rowcolor{lightblue}
Average & 51.9 & 36.4 & 34.2 & \textbf{56.7} & +4.8 \\
\midrule
\multicolumn{6}{l}{\textit{\textbf{GPT-OSS-20B}}} \\
\midrule
Biology & 47.4 & 57.9 & 63.2 & \textbf{84.2} & +36.8 \\
Physics & 62.8 & 68.6 & 66.3 & \textbf{80.2} & +17.4 \\
Chemistry & 47.3 & 34.4 & 41.9 & \textbf{74.2} & +26.9 \\
\rowcolor{lightblue}
Average & 52.5 & 53.6 & 57.1 & \textbf{79.5} & +27.1 \\
\bottomrule
\end{tabular}%
}
\vspace{5pt}
\caption{Performance (accuracy) comparison on the GPQA Diamond dataset across science domains. All values are reported in percentage (\%). Best results are \textbf{bolded}.}
\label{tab:gpqa_results}
\end{table}

\begin{table}[!t]
\centering
\resizebox{\columnwidth}{!}{
\begin{tabular}{lccccc}
\toprule
\textbf{Dataset} & \textbf{Orig.} & \textbf{RICE} & \textbf{SFT} & \textbf{DSMoE} & \textbf{+/-} \\
\midrule
\multicolumn{6}{l}{\textit{\textbf{Qwen3-30B-Instruct}}} \\ 
\midrule
AIME 24 & 56.7 & 63.3 & 63.3 & \textbf{70.0} & +13.3 \\
AIME 25 & 46.7 & 46.7 & 50.0 & \textbf{60.0} & +13.3 \\
\midrule
\multicolumn{6}{l}{\textit{\textbf{GPT-OSS-20B}}} \\ 
\midrule
AIME 24 & 50.0 & 56.7 & 56.7 & \textbf{77.3} & +27.3 \\
AIME 25 & 66.3 & 46.7 & 60.0 & \textbf{78.6} & +12.3 \\
\bottomrule
\end{tabular}%
}
\vspace{5pt}
\caption{Accuracy comparison on Math benchmarks (AIME 24 \& 25). All values are in percentages (\%). Best results are highlighted in \textbf{bold}.}
\label{tab:aime_results}
\end{table}

\subsection{Cost Analysis}

\noindent \textbf{One-time Cost.}\;
We analyze the computational cost of identifying domain-specific experts. For DSMoE, the one-time identification cost scales linearly with the number of samples in a domain, yielding a time complexity of $O(L)$ forward passes, where $L$ denotes the number of domain-specific samples. In contrast, the RICE baseline incurs a substantially higher cost of $O(L \times M)$ forward passes, where $M$ is the number of generated thinking tokens per sample. Since $M \gg 1$ in practice, this results in orders-of-magnitude higher computational overhead. Consequently, DSMoE is significantly more efficient than RICE for one-time expert identification.


\noindent \textbf{Inference Cost.}\; 
DSMoE uses a router weight steering approach, where the steered expert weights are computed once and stored for future inference. As a result, DSMoE maintains the same inference cost as the original MoE-based LLMs. Interestingly, DSMoE can produce answers using fewer thinking tokens, as illustrated in Table~\ref{tab:qualitative_math}, making MoE-based models more efficient during inference. RICE applies steering to the router scores on a per-sample basis, which cannot be precomputed. Consequently, RICE incurs higher inference cost compared to the original models.

\subsection{Ablation Studies}


To determine the optimal number of domain-specific experts, we conduct an ablation study on the number of activated domain-specific experts ($K$) using GPT-OSS-20B on the Biology domain, with results reported in Table~\ref{tab:ablation_k}. When $K=0$, corresponding to the original model without domain-specific routing, performance is lower than the best DSMoE configurations. As $K$ increases, performance initially improves and reaches its peak at $K=20$, indicating an effective balance between expert specialization and routing diversity. Beyond this point, further increases in $K$ yield diminishing returns. In practice, we find that setting the number of domain-specific experts to approximately 1\% of the total expert count often serves as an optimal hyperparameter.

\begin{table}[!t]
\centering
\begin{tabular}{cllc}
\toprule
\textbf{$K$} & \textbf{Domain} & \textbf{Method} & \textbf{GPT-OSS-20B} \\
\midrule
0 & \multirow{6}{*}{Biology} & Original & 76.4 \\
\cmidrule{3-4} 
5 & & \multirow{5}{*}{DSMoE} & 73.6 \\
10 & & & 73.3 \\
20 & & & \textbf{78.7} \\
30 & & & 77.4 \\
50 & & & 74.1 \\
\bottomrule
\end{tabular}%
\vspace{5pt}
\caption{Ablation study on the number of domain-specific experts ($K$) on the MMLU-Pro dataset. The horizontal line separates the baseline (Original) from the steering method (DSMoE). The best performance (accuracy) is achieved at $K=20$.}
\label{tab:ablation_k}
\end{table}


Table~\ref{tab:ablation_alpha} analyzes the effect of the steering coefficient ($\alpha$) on DSMoE performance using GPT-OSS-20B in the Biology domain with $K=20$. Compared to the original MoE-based LLMs, moderate steering improves performance, with optimal results achieved at $\alpha=5.0$. As $\alpha$ increases from lower values, performance gradually improves and reaches its peak at $\alpha=5.0$, before declining at higher values. This trend suggests that excessively small or large steering coefficients can lead to suboptimal routing behavior, either by providing insufficient guidance or by overly constraining expert selection. In practice, we observe that steering coefficients in the range of $[2.0, 5.0]$ consistently yield positive results across different configurations.

\begin{table}[!t]
\centering
\begin{tabular}{cllc}
\toprule
\textbf{$\alpha$} & \textbf{Domain} & \textbf{Method} & \textbf{GPT-OSS-20B} \\
\midrule
-- & \multirow{6}{*}{Biology} & Original & 76.4 \\
\cmidrule{3-4} 
0.1 & & \multirow{5}{*}{DSMoE} & 77.0 \\
0.5 & & & 76.3 \\
5.0 & & & \textbf{78.7} \\
10.0 & & & 77.0 \\
50.0 & & & 68.0 \\
\bottomrule
\end{tabular}%
\vspace{5pt}
\caption{Ablation study on the steering coefficient ($\alpha$) on the MMLU-Pro dataset. The Original method represents the baseline without steering. The best performance (accuracy) is observed at $\alpha=5.0$.}
\label{tab:ablation_alpha}
\end{table}

\section{Conclusion}
\label{sec:clus}

This paper addressed the existence of domain-specific experts in MoE-based LLMs. Following a comprehensive analysis of models up to 120B parameters, we confirmed that distinct experts align with specific domains. We subsequently proposed \textbf{Domain Steering Mixture of Experts (DSMoE)}, a training-free, zero-overhead framework. Extensive experiments confirm that DSMoE surpasses strong baselines, including Supervised Fine-Tuning, delivering consistent performance gains across both target and non-target domains. These findings suggest that exploiting intrinsic expert specialization is a highly efficient alternative to traditional fine-tuning.

\section*{Limitations}

This study focuses on enhancing the efficiency and effectiveness of MoE-based Large Language Models through a training-free approach. While the results are promising, our experiments were constrained by computational resources, limiting the evaluation to medium-scale datasets and models up to GPT-OSS-120B. Future work should assess the scalability of DSMoE beyond 400B parameters and benchmark it against other Large Reasoning Models, such as DeepSeek-R1.

\section*{Ethics Statement}
Despite the encouraging results, inference with large-scale LLMs remains highly resource-intensive, necessitating careful management of computational costs and environmental impact. Additionally, our study relies on web-sourced data, which may contain inherent gender and racial biases; future work should explore mitigation strategies to address these concerns. Finally, while this work represents a meaningful step toward advancing LLM development, it also underscores the importance of implementing robust safeguards to prevent potential misuse in harmful applications.



\bibliography{acl_latex}

@inproceedings{du_glam_2022,
	series = {Proceedings of {Machine} {Learning} {Research}},
	title = {{GLaM}: {Efficient} {Scaling} of {Language} {Models} with {Mixture}-of-{Experts}},
	volume = {162},
	url = {https://proceedings.mlr.press/v162/du22c.html},
	booktitle = {Proceedings of the 39th {International} {Conference} on {Machine} {Learning}},
	publisher = {PMLR},
	author = {Du, Nan and Huang, Yanping and Dai, Andrew M and Tong, Simon and Lepikhin, Dmitry and Xu, Yuanzhong and Krikun, Maxim and Zhou, Yanqi and Yu, Adams Wei and Firat, Orhan and Zoph, Barret and Fedus, Liam and Bosma, Maarten P and Zhou, Zongwei and Wang, Tao and Wang, Emma and Webster, Kellie and Pellat, Marie and Robinson, Kevin and Meier-Hellstern, Kathleen and Duke, Toju and Dixon, Lucas and Zhang, Kun and Le, Quoc and Wu, Yonghui and Chen, Zhifeng and Cui, Claire},
	editor = {Chaudhuri, Kamalika and Jegelka, Stefanie and Song, Le and Szepesvari, Csaba and Niu, Gang and Sabato, Sivan},
	month = jul,
	year = {2022},
	pages = {5547--5569},
}

@article{fedus_switch_2022,
	title = {Switch {Transformers}: {Scaling} to {Trillion} {Parameter} {Models} with {Simple} and {Efficient} {Sparsity}},
	volume = {23},
	url = {http://jmlr.org/papers/v23/21-0998.html},
	number = {120},
	journal = {Journal of Machine Learning Research},
	author = {Fedus, William and Zoph, Barret and Shazeer, Noam},
	year = {2022},
	pages = {1--39},
}

@inproceedings{devlin-etal-2019-bert,
    title = "{BERT}: Pre-training of Deep Bidirectional Transformers for Language Understanding",
    author = "Devlin, Jacob  and
      Chang, Ming-Wei  and
      Lee, Kenton  and
      Toutanova, Kristina",
    editor = "Burstein, Jill  and
      Doran, Christy  and
      Solorio, Thamar",
    booktitle = "Proceedings of the 2019 Conference of the North {A}merican Chapter of the Association for Computational Linguistics: Human Language Technologies, Volume 1 (Long and Short Papers)",
    month = jun,
    year = "2019",
    address = "Minneapolis, Minnesota",
    publisher = "Association for Computational Linguistics",
    url = "https://aclanthology.org/N19-1423",
    doi = "10.18653/v1/N19-1423",
    pages = "4171--4186",
}

@misc{shazeer2017outrageously,
      title={Outrageously Large Neural Networks: The Sparsely-Gated Mixture-of-Experts Layer}, 
      author={Noam Shazeer and Azalia Mirhoseini and Krzysztof Maziarz and Andy Davis and Quoc Le and Geoffrey Hinton and Jeff Dean},
      year={2017},
      eprint={1701.06538},
      archivePrefix={arXiv},
      primaryClass={cs.LG}
}

@inproceedings{NEURIPS2021_48237d9f,
 author = {Riquelme, Carlos and Puigcerver, Joan and Mustafa, Basil and Neumann, Maxim and Jenatton, Rodolphe and Susano Pinto, Andr\'{e} and Keysers, Daniel and Houlsby, Neil},
 booktitle = {Advances in Neural Information Processing Systems},
 editor = {M. Ranzato and A. Beygelzimer and Y. Dauphin and P.S. Liang and J. Wortman Vaughan},
 pages = {8583--8595},
 publisher = {Curran Associates, Inc.},
 title = {Scaling Vision with Sparse Mixture of Experts},
 url = {https://proceedings.neurips.cc/paper_files/paper/2021/file/48237d9f2dea8c74c2a72126cf63d933-Paper.pdf},
 volume = {34},
 year = {2021}
}

@misc{chi2022representation,
      title={On the Representation Collapse of Sparse Mixture of Experts}, 
      author={Zewen Chi and Li Dong and Shaohan Huang and Damai Dai and Shuming Ma and Barun Patra and Saksham Singhal and Payal Bajaj and Xia Song and Xian-Ling Mao and Heyan Huang and Furu Wei},
      year={2022},
      eprint={2204.09179},
      archivePrefix={arXiv},
      primaryClass={cs.CL}
}

@misc{do2023hyperrouter,
      title={HyperRouter: Towards Efficient Training and Inference of Sparse Mixture of Experts}, 
      author={Giang Do and Khiem Le and Quang Pham and TrungTin Nguyen and Thanh-Nam Doan and Bint T. Nguyen and Chenghao Liu and Savitha Ramasamy and Xiaoli Li and Steven Hoi},
      year={2023},
      eprint={2312.07035},
      archivePrefix={arXiv},
      primaryClass={cs.LG}
}

@misc{chen2023sparse,
      title={Sparse MoE as the New Dropout: Scaling Dense and Self-Slimmable Transformers}, 
      author={Tianlong Chen and Zhenyu Zhang and Ajay Jaiswal and Shiwei Liu and Zhangyang Wang},
      year={2023},
      eprint={2303.01610},
      archivePrefix={arXiv},
      primaryClass={cs.LG}
}

@misc{dai2022stablemoe,
      title={StableMoE: Stable Routing Strategy for Mixture of Experts}, 
      author={Damai Dai and Li Dong and Shuming Ma and Bo Zheng and Zhifang Sui and Baobao Chang and Furu Wei},
      year={2022},
      eprint={2204.08396},
      archivePrefix={arXiv},
      primaryClass={cs.LG}
}

@inproceedings{NEURIPS2022_2f00ecd7,
 author = {Zhou, Yanqi and Lei, Tao and Liu, Hanxiao and Du, Nan and Huang, Yanping and Zhao, Vincent and Dai, Andrew M and Chen, zhifeng and Le, Quoc V and Laudon, James},
 booktitle = {Advances in Neural Information Processing Systems},
 editor = {S. Koyejo and S. Mohamed and A. Agarwal and D. Belgrave and K. Cho and A. Oh},
 pages = {7103--7114},
 publisher = {Curran Associates, Inc.},
 title = {Mixture-of-Experts with Expert Choice Routing},
 url = {https://proceedings.neurips.cc/paper_files/paper/2022/file/2f00ecd787b432c1d36f3de9800728eb-Paper-Conference.pdf},
 volume = {35},
 year = {2022}
}

@ARTICLE{jacobs1991,
  author={Jacobs, Robert A. and Jordan, Michael I. and Nowlan, Steven J. and Hinton, Geoffrey E.},
  journal={Neural Computation}, 
  title={Adaptive Mixtures of Local Experts}, 
  year={1991},
  volume={3},
  number={1},
  pages={79-87},
  doi={10.1162/neco.1991.3.1.79}}

@article{jordan1994,
author = {Jordan, Michael and Jacobs, Robert},
year = {1994},
month = {01},
pages = {181-},
title = {Hierarchical mixtures of experts and the},
volume = {6},
journal = {Neural computation}
}

@misc{riquelme2021scalingvisionsparsemixture,
      title={Scaling Vision with Sparse Mixture of Experts}, 
      author={Carlos Riquelme and Joan Puigcerver and Basil Mustafa and Maxim Neumann and Rodolphe Jenatton and André Susano Pinto and Daniel Keysers and Neil Houlsby},
      year={2021},
      eprint={2106.05974},
      archivePrefix={arXiv},
      primaryClass={cs.CV},
      url={https://arxiv.org/abs/2106.05974}, 
}

@inproceedings{shen-etal-2023-scaling,
    title = "Scaling Vision-Language Models with Sparse Mixture of Experts",
    author = "Shen, Sheng  and
      Yao, Zhewei  and
      Li, Chunyuan  and
      Darrell, Trevor  and
      Keutzer, Kurt  and
      He, Yuxiong",
    editor = "Bouamor, Houda  and
      Pino, Juan  and
      Bali, Kalika",
    booktitle = "Findings of the Association for Computational Linguistics: EMNLP 2023",
    month = dec,
    year = "2023",
    address = "Singapore",
    publisher = "Association for Computational Linguistics",
    url = "https://aclanthology.org/2023.findings-emnlp.758",
    doi = "10.18653/v1/2023.findings-emnlp.758",
    pages = "11329--11344",
}

@misc{do2024simsmoesolvingrepresentationalcollapse,
      title={SimSMoE: Solving Representational Collapse via Similarity Measure}, 
      author={Giang Do and Hung Le and Truyen Tran},
      year={2024},
      eprint={2406.15883},
      archivePrefix={arXiv},
      primaryClass={cs.CL},
      url={https://arxiv.org/abs/2406.15883}, 
}

@article{article,
author = {Friedman, Nir and Geiger, Dan and Goldszmidt, Moises},
year = {1997},
month = {11},
pages = {131-163},
title = {Bayesian Network Classifiers},
volume = {29},
journal = {Machine Learning},
doi = {10.1023/A:1007465528199}
}

@misc{muennighoff2024olmoeopenmixtureofexpertslanguage,
      title={OLMoE: Open Mixture-of-Experts Language Models}, 
      author={Niklas Muennighoff and Luca Soldaini and Dirk Groeneveld and Kyle Lo and Jacob Morrison and Sewon Min and Weijia Shi and Pete Walsh and Oyvind Tafjord and Nathan Lambert and Yuling Gu and Shane Arora and Akshita Bhagia and Dustin Schwenk and David Wadden and Alexander Wettig and Binyuan Hui and Tim Dettmers and Douwe Kiela and Ali Farhadi and Noah A. Smith and Pang Wei Koh and Amanpreet Singh and Hannaneh Hajishirzi},
      year={2024},
      eprint={2409.02060},
      archivePrefix={arXiv},
      primaryClass={cs.CL},
      url={https://arxiv.org/abs/2409.02060}, 
}

@misc{dai2024deepseekmoeultimateexpertspecialization,
      title={DeepSeekMoE: Towards Ultimate Expert Specialization in Mixture-of-Experts Language Models}, 
      author={Damai Dai and Chengqi Deng and Chenggang Zhao and R. X. Xu and Huazuo Gao and Deli Chen and Jiashi Li and Wangding Zeng and Xingkai Yu and Y. Wu and Zhenda Xie and Y. K. Li and Panpan Huang and Fuli Luo and Chong Ruan and Zhifang Sui and Wenfeng Liang},
      year={2024},
      eprint={2401.06066},
      archivePrefix={arXiv},
      primaryClass={cs.CL},
      url={https://arxiv.org/abs/2401.06066}, 
}

@misc{qwen_moe,
    title = {Qwen1.5-MoE: Matching 7B Model Performance with 1/3 Activated Parameters"},
    url = {https://qwenlm.github.io/blog/qwen-moe/},
    author = {Qwen Team},
    month = {February},
    year = {2024}
}

@misc{jiang2024mixtralexperts,
      title={Mixtral of Experts}, 
      author={Albert Q. Jiang and Alexandre Sablayrolles and Antoine Roux and Arthur Mensch and Blanche Savary and Chris Bamford and Devendra Singh Chaplot and Diego de las Casas and Emma Bou Hanna and Florian Bressand and Gianna Lengyel and Guillaume Bour and Guillaume Lample and Lélio Renard Lavaud and Lucile Saulnier and Marie-Anne Lachaux and Pierre Stock and Sandeep Subramanian and Sophia Yang and Szymon Antoniak and Teven Le Scao and Théophile Gervet and Thibaut Lavril and Thomas Wang and Timothée Lacroix and William El Sayed},
      year={2024},
      eprint={2401.04088},
      archivePrefix={arXiv},
      primaryClass={cs.LG},
      url={https://arxiv.org/abs/2401.04088}, 
}

@inproceedings{NEURIPS2020_1457c0d6,
 author = {Brown, Tom and Mann, Benjamin and Ryder, Nick and Subbiah, Melanie and Kaplan, Jared D and Dhariwal, Prafulla and Neelakantan, Arvind and Shyam, Pranav and Sastry, Girish and Askell, Amanda and Agarwal, Sandhini and Herbert-Voss, Ariel and Krueger, Gretchen and Henighan, Tom and Child, Rewon and Ramesh, Aditya and Ziegler, Daniel and Wu, Jeffrey and Winter, Clemens and Hesse, Chris and Chen, Mark and Sigler, Eric and Litwin, Mateusz and Gray, Scott and Chess, Benjamin and Clark, Jack and Berner, Christopher and McCandlish, Sam and Radford, Alec and Sutskever, Ilya and Amodei, Dario},
 booktitle = {Advances in Neural Information Processing Systems},
 editor = {H. Larochelle and M. Ranzato and R. Hadsell and M.F. Balcan and H. Lin},
 pages = {1877--1901},
 publisher = {Curran Associates, Inc.},
 title = {Language Models are Few-Shot Learners},
 url = {https://proceedings.neurips.cc/paper_files/paper/2020/file/1457c0d6bfcb4967418bfb8ac142f64a-Paper.pdf},
 volume = {33},
 year = {2020}
}

@inproceedings{zhan-etal-2024-anygpt,
    title = "{A}ny{GPT}: Unified Multimodal {LLM} with Discrete Sequence Modeling",
    author = "Zhan, Jun  and
      Dai, Junqi  and
      Ye, Jiasheng  and
      Zhou, Yunhua  and
      Zhang, Dong  and
      Liu, Zhigeng  and
      Zhang, Xin  and
      Yuan, Ruibin  and
      Zhang, Ge  and
      Li, Linyang  and
      Yan, Hang  and
      Fu, Jie  and
      Gui, Tao  and
      Sun, Tianxiang  and
      Jiang, Yu-Gang  and
      Qiu, Xipeng",
    editor = "Ku, Lun-Wei  and
      Martins, Andre  and
      Srikumar, Vivek",
    booktitle = "Proceedings of the 62nd Annual Meeting of the Association for Computational Linguistics (Volume 1: Long Papers)",
    month = aug,
    year = "2024",
    address = "Bangkok, Thailand",
    publisher = "Association for Computational Linguistics",
    url = "https://aclanthology.org/2024.acl-long.521/",
    doi = "10.18653/v1/2024.acl-long.521",
    pages = "9637--9662"
}

@ARTICLE{10887014,
  author={Li, Yunxin and Jiang, Shenyuan and Hu, Baotian and Wang, Longyue and Zhong, Wanqi and Luo, Wenhan and Ma, Lin and Zhang, Min},
  journal={IEEE Transactions on Pattern Analysis and Machine Intelligence}, 
  title={Uni-MoE: Scaling Unified Multimodal LLMs With Mixture of Experts}, 
  year={2025},
  volume={47},
  number={5},
  pages={3424-3439},
  doi={10.1109/TPAMI.2025.3532688}}

@misc{kaplan2020scalinglawsneurallanguage,
      title={Scaling Laws for Neural Language Models}, 
      author={Jared Kaplan and Sam McCandlish and Tom Henighan and Tom B. Brown and Benjamin Chess and Rewon Child and Scott Gray and Alec Radford and Jeffrey Wu and Dario Amodei},
      year={2020},
      eprint={2001.08361},
      archivePrefix={arXiv},
      primaryClass={cs.LG},
      url={https://arxiv.org/abs/2001.08361}, 
}

@misc{krishnamurthy2023improvingexpertspecializationmixture,
      title={Improving Expert Specialization in Mixture of Experts}, 
      author={Yamuna Krishnamurthy and Chris Watkins and Thomas Gaertner},
      year={2023},
      eprint={2302.14703},
      archivePrefix={arXiv},
      primaryClass={cs.LG},
      url={https://arxiv.org/abs/2302.14703}, 
}

@misc{wang2024letexpertsticklast,
      title={Let the Expert Stick to His Last: Expert-Specialized Fine-Tuning for Sparse Architectural Large Language Models}, 
      author={Zihan Wang and Deli Chen and Damai Dai and Runxin Xu and Zhuoshu Li and Y. Wu},
      year={2024},
      eprint={2407.01906},
      archivePrefix={arXiv},
      primaryClass={cs.CL},
      url={https://arxiv.org/abs/2407.01906}, 
}

@misc{openai2025gptoss120bgptoss20bmodel,
      title={gpt-oss-120b \& gpt-oss-20b Model Card}, 
      author={OpenAI and : and Sandhini Agarwal and Lama Ahmad and Jason Ai and Sam Altman and Andy Applebaum and Edwin Arbus and Rahul K. Arora and Yu Bai and Bowen Baker and Haiming Bao and Boaz Barak and Ally Bennett and Tyler Bertao and Nivedita Brett and Eugene Brevdo and Greg Brockman and Sebastien Bubeck and Che Chang and Kai Chen and Mark Chen and Enoch Cheung and Aidan Clark and Dan Cook and Marat Dukhan and Casey Dvorak and Kevin Fives and Vlad Fomenko and Timur Garipov and Kristian Georgiev and Mia Glaese and Tarun Gogineni and Adam Goucher and Lukas Gross and Katia Gil Guzman and John Hallman and Jackie Hehir and Johannes Heidecke and Alec Helyar and Haitang Hu and Romain Huet and Jacob Huh and Saachi Jain and Zach Johnson and Chris Koch and Irina Kofman and Dominik Kundel and Jason Kwon and Volodymyr Kyrylov and Elaine Ya Le and Guillaume Leclerc and James Park Lennon and Scott Lessans and Mario Lezcano-Casado and Yuanzhi Li and Zhuohan Li and Ji Lin and Jordan Liss and Lily and Liu and Jiancheng Liu and Kevin Lu and Chris Lu and Zoran Martinovic and Lindsay McCallum and Josh McGrath and Scott McKinney and Aidan McLaughlin and Song Mei and Steve Mostovoy and Tong Mu and Gideon Myles and Alexander Neitz and Alex Nichol and Jakub Pachocki and Alex Paino and Dana Palmie and Ashley Pantuliano and Giambattista Parascandolo and Jongsoo Park and Leher Pathak and Carolina Paz and Ludovic Peran and Dmitry Pimenov and Michelle Pokrass and Elizabeth Proehl and Huida Qiu and Gaby Raila and Filippo Raso and Hongyu Ren and Kimmy Richardson and David Robinson and Bob Rotsted and Hadi Salman and Suvansh Sanjeev and Max Schwarzer and D. Sculley and Harshit Sikchi and Kendal Simon and Karan Singhal and Yang Song and Dane Stuckey and Zhiqing Sun and Philippe Tillet and Sam Toizer and Foivos Tsimpourlas and Nikhil Vyas and Eric Wallace and Xin Wang and Miles Wang and Olivia Watkins and Kevin Weil and Amy Wendling and Kevin Whinnery and Cedric Whitney and Hannah Wong and Lin Yang and Yu Yang and Michihiro Yasunaga and Kristen Ying and Wojciech Zaremba and Wenting Zhan and Cyril Zhang and Brian Zhang and Eddie Zhang and Shengjia Zhao},
      year={2025},
      eprint={2508.10925},
      archivePrefix={arXiv},
      primaryClass={cs.CL},
      url={https://arxiv.org/abs/2508.10925}, 
}

@misc{li2025slimmoestructuredcompressionlarge,
      title={SlimMoE: Structured Compression of Large MoE Models via Expert Slimming and Distillation}, 
      author={Zichong Li and Chen Liang and Zixuan Zhang and Ilgee Hong and Young Jin Kim and Weizhu Chen and Tuo Zhao},
      year={2025},
      eprint={2506.18349},
      archivePrefix={arXiv},
      primaryClass={cs.LG},
      url={https://arxiv.org/abs/2506.18349}, 
}

@misc{yang2025qwen3technicalreport,
      title={Qwen3 Technical Report}, 
      author={An Yang and Anfeng Li and Baosong Yang and Beichen Zhang and Binyuan Hui and Bo Zheng and Bowen Yu and Chang Gao and Chengen Huang and Chenxu Lv and Chujie Zheng and Dayiheng Liu and Fan Zhou and Fei Huang and Feng Hu and Hao Ge and Haoran Wei and Huan Lin and Jialong Tang and Jian Yang and Jianhong Tu and Jianwei Zhang and Jianxin Yang and Jiaxi Yang and Jing Zhou and Jingren Zhou and Junyang Lin and Kai Dang and Keqin Bao and Kexin Yang and Le Yu and Lianghao Deng and Mei Li and Mingfeng Xue and Mingze Li and Pei Zhang and Peng Wang and Qin Zhu and Rui Men and Ruize Gao and Shixuan Liu and Shuang Luo and Tianhao Li and Tianyi Tang and Wenbiao Yin and Xingzhang Ren and Xinyu Wang and Xinyu Zhang and Xuancheng Ren and Yang Fan and Yang Su and Yichang Zhang and Yinger Zhang and Yu Wan and Yuqiong Liu and Zekun Wang and Zeyu Cui and Zhenru Zhang and Zhipeng Zhou and Zihan Qiu},
      year={2025},
      eprint={2505.09388},
      archivePrefix={arXiv},
      primaryClass={cs.CL},
      url={https://arxiv.org/abs/2505.09388}, 
}

@misc{wang2024mmluprorobustchallengingmultitask,
      title={MMLU-Pro: A More Robust and Challenging Multi-Task Language Understanding Benchmark}, 
      author={Yubo Wang and Xueguang Ma and Ge Zhang and Yuansheng Ni and Abhranil Chandra and Shiguang Guo and Weiming Ren and Aaran Arulraj and Xuan He and Ziyan Jiang and Tianle Li and Max Ku and Kai Wang and Alex Zhuang and Rongqi Fan and Xiang Yue and Wenhu Chen},
      year={2024},
      eprint={2406.01574},
      archivePrefix={arXiv},
      primaryClass={cs.CL},
      url={https://arxiv.org/abs/2406.01574}, 
}

@misc{rein2023gpqagraduatelevelgoogleproofqa,
      title={GPQA: A Graduate-Level Google-Proof Q\&A Benchmark}, 
      author={David Rein and Betty Li Hou and Asa Cooper Stickland and Jackson Petty and Richard Yuanzhe Pang and Julien Dirani and Julian Michael and Samuel R. Bowman},
      year={2023},
      eprint={2311.12022},
      archivePrefix={arXiv},
      primaryClass={cs.AI},
      url={https://arxiv.org/abs/2311.12022}, 
}

@inproceedings{NEURIPS2024_5eeb693f,
 author = {Oldfield, James and Georgopoulos, Markos and Chrysos, Grigorios G. and Tzelepis, Christos and Panagakis, Yannis and Nicolaou, Mihalis A. and Deng, Jiankang and Patras, Ioannis},
 booktitle = {Advances in Neural Information Processing Systems},
 doi = {10.52202/079017-1680},
 editor = {A. Globerson and L. Mackey and D. Belgrave and A. Fan and U. Paquet and J. Tomczak and C. Zhang},
 pages = {53022--53063},
 publisher = {Curran Associates, Inc.},
 title = {Multilinear Mixture of Experts: Scalable Expert Specialization through Factorization},
 url = {https://proceedings.neurips.cc/paper_files/paper/2024/file/5eeb693f46d753e5fe24c97212c22bd2-Paper-Conference.pdf},
 volume = {37},
 year = {2024}
}

@misc{guo2025advancingexpertspecializationbetter,
      title={Advancing Expert Specialization for Better MoE}, 
      author={Hongcan Guo and Haolang Lu and Guoshun Nan and Bolun Chu and Jialin Zhuang and Yuan Yang and Wenhao Che and Sicong Leng and Qimei Cui and Xudong Jiang},
      year={2025},
      eprint={2505.22323},
      archivePrefix={arXiv},
      primaryClass={cs.CL},
      url={https://arxiv.org/abs/2505.22323}, 
}

@misc{ismail2023interpretablemixtureexperts,
      title={Interpretable Mixture of Experts}, 
      author={Aya Abdelsalam Ismail and Sercan Ö. Arik and Jinsung Yoon and Ankur Taly and Soheil Feizi and Tomas Pfister},
      year={2023},
      eprint={2206.02107},
      archivePrefix={arXiv},
      primaryClass={cs.LG},
      url={https://arxiv.org/abs/2206.02107}, 
}

@misc{nikolic2025exploringexpertspecializationunsupervised,
      title={Exploring Expert Specialization through Unsupervised Training in Sparse Mixture of Experts}, 
      author={Strahinja Nikolic and Ilker Oguz and Demetri Psaltis},
      year={2025},
      eprint={2509.10025},
      archivePrefix={arXiv},
      primaryClass={cs.LG},
      url={https://arxiv.org/abs/2509.10025}, 
}

@inproceedings{
chaudhari2025moe,
title={MoE Lens - An Expert Is All You Need},
author={Marmik Chaudhari and Idhant Gulati and Nishkal Hundia and Pranav Karra and Shivam Raval},
booktitle={Sparsity in LLMs (SLLM): Deep Dive into Mixture of Experts, Quantization, Hardware, and Inference},
year={2025},
url={https://openreview.net/forum?id=GS4WXncwSF}
}

@misc{turner2024steeringlanguagemodelsactivation,
      title={Steering Language Models With Activation Engineering}, 
      author={Alexander Matt Turner and Lisa Thiergart and Gavin Leech and David Udell and Juan J. Vazquez and Ulisse Mini and Monte MacDiarmid},
      year={2024},
      eprint={2308.10248},
      archivePrefix={arXiv},
      primaryClass={cs.CL},
      url={https://arxiv.org/abs/2308.10248}, 
}

@inproceedings{rimsky-etal-2024-steering,
    title = "Steering Llama 2 via Contrastive Activation Addition",
    author = "Rimsky, Nina  and
      Gabrieli, Nick  and
      Schulz, Julian  and
      Tong, Meg  and
      Hubinger, Evan  and
      Turner, Alexander",
    editor = "Ku, Lun-Wei  and
      Martins, Andre  and
      Srikumar, Vivek",
    booktitle = "Proceedings of the 62nd Annual Meeting of the Association for Computational Linguistics (Volume 1: Long Papers)",
    month = aug,
    year = "2024",
    address = "Bangkok, Thailand",
    publisher = "Association for Computational Linguistics",
    url = "https://aclanthology.org/2024.acl-long.828/",
    doi = "10.18653/v1/2024.acl-long.828",
    pages = "15504--15522"
}

@inproceedings{
wang2025two,
title={Two Experts Are All You Need for Steering Thinking: Reinforcing Cognitive Effort in MoE Reasoning Models Without Additional Training},
author={Mengru Wang and Xingyu Chen and Yue Wang and Zhiwei He and Jiahao Xu and Tian Liang and Qiuzhi Liu and Yunzhi Yao and Wenxuan Wang and Ruotian Ma and Haitao Mi and Ningyu Zhang and Zhaopeng Tu and Xiaolong Li and Dong Yu},
booktitle={The Thirty-ninth Annual Conference on Neural Information Processing Systems},
year={2025},
url={https://openreview.net/forum?id=x7fCiuCCAu}
}

@misc{deepseekai2025deepseekr1incentivizingreasoningcapability,
      title={DeepSeek-R1: Incentivizing Reasoning Capability in LLMs via Reinforcement Learning}, 
      author={DeepSeek-AI and Daya Guo and Dejian Yang and Haowei Zhang and Junxiao Song and Ruoyu Zhang and Runxin Xu and Qihao Zhu and Shirong Ma and Peiyi Wang and Xiao Bi and Xiaokang Zhang and Xingkai Yu and Yu Wu and Z. F. Wu and Zhibin Gou and Zhihong Shao and Zhuoshu Li and Ziyi Gao and Aixin Liu and Bing Xue and Bingxuan Wang and Bochao Wu and Bei Feng and Chengda Lu and Chenggang Zhao and Chengqi Deng and Chenyu Zhang and Chong Ruan and Damai Dai and Deli Chen and Dongjie Ji and Erhang Li and Fangyun Lin and Fucong Dai and Fuli Luo and Guangbo Hao and Guanting Chen and Guowei Li and H. Zhang and Han Bao and Hanwei Xu and Haocheng Wang and Honghui Ding and Huajian Xin and Huazuo Gao and Hui Qu and Hui Li and Jianzhong Guo and Jiashi Li and Jiawei Wang and Jingchang Chen and Jingyang Yuan and Junjie Qiu and Junlong Li and J. L. Cai and Jiaqi Ni and Jian Liang and Jin Chen and Kai Dong and Kai Hu and Kaige Gao and Kang Guan and Kexin Huang and Kuai Yu and Lean Wang and Lecong Zhang and Liang Zhao and Litong Wang and Liyue Zhang and Lei Xu and Leyi Xia and Mingchuan Zhang and Minghua Zhang and Minghui Tang and Meng Li and Miaojun Wang and Mingming Li and Ning Tian and Panpan Huang and Peng Zhang and Qiancheng Wang and Qinyu Chen and Qiushi Du and Ruiqi Ge and Ruisong Zhang and Ruizhe Pan and Runji Wang and R. J. Chen and R. L. Jin and Ruyi Chen and Shanghao Lu and Shangyan Zhou and Shanhuang Chen and Shengfeng Ye and Shiyu Wang and Shuiping Yu and Shunfeng Zhou and Shuting Pan and S. S. Li and Shuang Zhou and Shaoqing Wu and Shengfeng Ye and Tao Yun and Tian Pei and Tianyu Sun and T. Wang and Wangding Zeng and Wanjia Zhao and Wen Liu and Wenfeng Liang and Wenjun Gao and Wenqin Yu and Wentao Zhang and W. L. Xiao and Wei An and Xiaodong Liu and Xiaohan Wang and Xiaokang Chen and Xiaotao Nie and Xin Cheng and Xin Liu and Xin Xie and Xingchao Liu and Xinyu Yang and Xinyuan Li and Xuecheng Su and Xuheng Lin and X. Q. Li and Xiangyue Jin and Xiaojin Shen and Xiaosha Chen and Xiaowen Sun and Xiaoxiang Wang and Xinnan Song and Xinyi Zhou and Xianzu Wang and Xinxia Shan and Y. K. Li and Y. Q. Wang and Y. X. Wei and Yang Zhang and Yanhong Xu and Yao Li and Yao Zhao and Yaofeng Sun and Yaohui Wang and Yi Yu and Yichao Zhang and Yifan Shi and Yiliang Xiong and Ying He and Yishi Piao and Yisong Wang and Yixuan Tan and Yiyang Ma and Yiyuan Liu and Yongqiang Guo and Yuan Ou and Yuduan Wang and Yue Gong and Yuheng Zou and Yujia He and Yunfan Xiong and Yuxiang Luo and Yuxiang You and Yuxuan Liu and Yuyang Zhou and Y. X. Zhu and Yanhong Xu and Yanping Huang and Yaohui Li and Yi Zheng and Yuchen Zhu and Yunxian Ma and Ying Tang and Yukun Zha and Yuting Yan and Z. Z. Ren and Zehui Ren and Zhangli Sha and Zhe Fu and Zhean Xu and Zhenda Xie and Zhengyan Zhang and Zhewen Hao and Zhicheng Ma and Zhigang Yan and Zhiyu Wu and Zihui Gu and Zijia Zhu and Zijun Liu and Zilin Li and Ziwei Xie and Ziyang Song and Zizheng Pan and Zhen Huang and Zhipeng Xu and Zhongyu Zhang and Zhen Zhang},
      year={2025},
      eprint={2501.12948},
      archivePrefix={arXiv},
      primaryClass={cs.CL},
      url={https://arxiv.org/abs/2501.12948}, 
}

@misc{qwen3technicalreport,
      title={Qwen3 Technical Report}, 
      author={Qwen Team},
      year={2025},
      eprint={2505.09388},
      archivePrefix={arXiv},
      primaryClass={cs.CL},
      url={https://arxiv.org/abs/2505.09388}, 
}

@INPROCEEDINGS{1716307,
  author={Cawley, G.C.},
  booktitle={The 2006 IEEE International Joint Conference on Neural Network Proceedings}, 
  title={Leave-One-Out Cross-Validation Based Model Selection Criteria for Weighted LS-SVMs}, 
  year={2006},
  volume={},
  number={},
  pages={1661-1668},
  doi={10.1109/IJCNN.2006.246634}}

@misc{shrikumar2019learningimportantfeaturespropagating,
      title={Learning Important Features Through Propagating Activation Differences}, 
      author={Avanti Shrikumar and Peyton Greenside and Anshul Kundaje},
      year={2019},
      eprint={1704.02685},
      archivePrefix={arXiv},
      primaryClass={cs.CV},
      url={https://arxiv.org/abs/1704.02685}, 
}

@misc{ancona2018betterunderstandinggradientbasedattribution,
      title={Towards better understanding of gradient-based attribution methods for Deep Neural Networks}, 
      author={Marco Ancona and Enea Ceolini and Cengiz Öztireli and Markus Gross},
      year={2018},
      eprint={1711.06104},
      archivePrefix={arXiv},
      primaryClass={cs.LG},
      url={https://arxiv.org/abs/1711.06104}, 
}

@misc{ernie2025technicalreport,
      title={ERNIE 4.5 Technical Report},
      author={Baidu-ERNIE-Team},
      year={2025},
      primaryClass={cs.CL},
      howpublished={\url{https://ernie.baidu.com/blog/publication/ERNIE_Technical_Report.pdf}}
}

@misc{wang2025expertsneedsteeringthinking,
      title={Two Experts Are All You Need for Steering Thinking: Reinforcing Cognitive Effort in MoE Reasoning Models Without Additional Training}, 
      author={Mengru Wang and Xingyu Chen and Yue Wang and Zhiwei He and Jiahao Xu and Tian Liang and Qiuzhi Liu and Yunzhi Yao and Wenxuan Wang and Ruotian Ma and Haitao Mi and Ningyu Zhang and Zhaopeng Tu and Xiaolong Li and Dong Yu},
      year={2025},
      eprint={2505.14681},
      archivePrefix={arXiv},
      primaryClass={cs.AI},
      url={https://arxiv.org/abs/2505.14681}, 
}

@misc{hu2021loralowrankadaptationlarge,
      title={LoRA: Low-Rank Adaptation of Large Language Models}, 
      author={Edward J. Hu and Yelong Shen and Phillip Wallis and Zeyuan Allen-Zhu and Yuanzhi Li and Shean Wang and Lu Wang and Weizhu Chen},
      year={2021},
      eprint={2106.09685},
      archivePrefix={arXiv},
      primaryClass={cs.CL},
      url={https://arxiv.org/abs/2106.09685}, 
}

@misc{maa_aime,
  author       = {{MAA Committees}},
  title        = {{AIME} Problems and Solutions},
  howpublished = {\url{https://artofproblemsolving.com/wiki/index.php/AIME_Problems_and_Solutions}},
  year         = {n.d.},
}

@inproceedings{kwon2023efficient,
  title={Efficient Memory Management for Large Language Model Serving with PagedAttention},
  author={Woosuk Kwon and Zhuohan Li and Siyuan Zhuang and Ying Sheng and Lianmin Zheng and Cody Hao Yu and Joseph E. Gonzalez and Hao Zhang and Ion Stoica},
  booktitle={Proceedings of the ACM SIGOPS 29th Symposium on Operating Systems Principles},
  year={2023}
}

@Misc{peft,
  title =        {{PEFT}: State-of-the-art Parameter-Efficient Fine-Tuning methods},
  author =       {Sourab Mangrulkar and Sylvain Gugger and Lysandre Debut and Younes Belkada and Sayak Paul and Benjamin Bossan and Marian Tietz},
  howpublished = {\url{https://github.com/huggingface/peft}},
  year =         {2022}
}

\appendix

\section{Appendix}
\label{sec:appendix}

This document is organized as follows: Appendix~\ref{app:add_anal} illustrate some further analysis of DSMoE. Appendix~\ref{app:add_imp} presents supplementary benchmarks descriptions, and Appendix~\ref{app:imp} describes the implementation details in full.

\subsection{Domain-specific Experts Analysis}
\label{app:add_anal}

\begin{figure*}[t]
    \centering
    \begin{subfigure}[b]{\textwidth}
        \includegraphics[width=\textwidth]{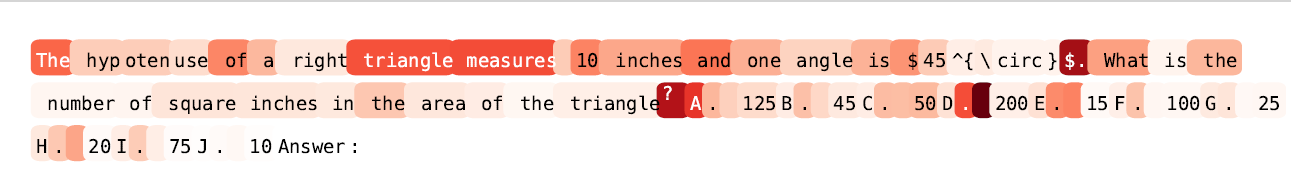}
        \caption{Sample 1}
        \label{fig:token1}
    \end{subfigure}
    
    \vspace{0.3cm}
    
    \begin{subfigure}[b]{\textwidth}
        \includegraphics[width=\textwidth]{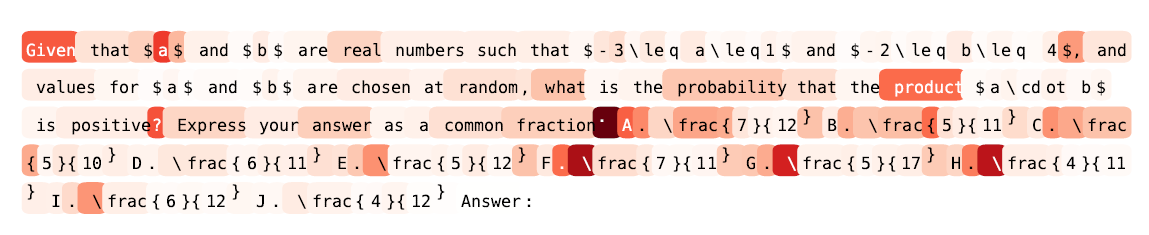}
        \caption{Sample 2}
        \label{fig:token2}
    \end{subfigure}
    
    \vspace{0.3cm}
    
    \begin{subfigure}[b]{\textwidth}
        \includegraphics[width=\textwidth]{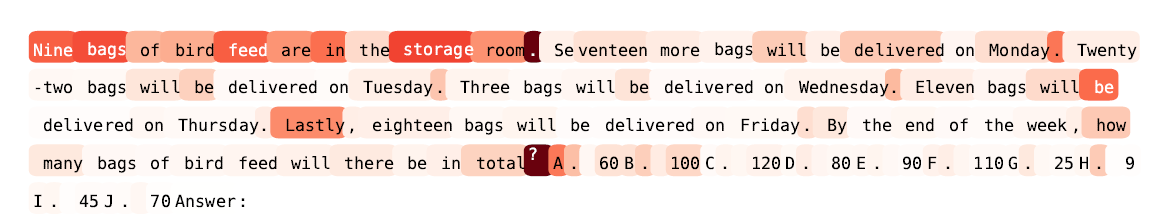}
        \caption{Sample 3}
        \label{fig:token3}
    \end{subfigure}
    
    \vspace{0.3cm}
    
    \begin{subfigure}[b]{\textwidth}
        \includegraphics[width=\textwidth]{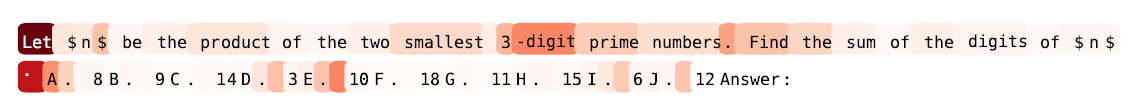}
        \caption{Sample 4}
        \label{fig:token4}
    \end{subfigure}
    
    \vspace{0.3cm}
    
    \begin{subfigure}[b]{\textwidth}
        \includegraphics[width=\textwidth]{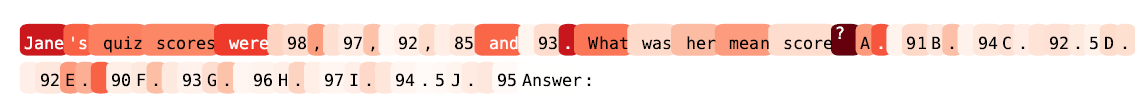}
        \caption{Sample 5}
        \label{fig:token5}
    \end{subfigure}
    
    \caption{Token ranking scores across five representative samples from the MMLU-Pro, mathematics domain for GPT-OSS-20B. Each row displays the importance distribution of tokens within a single question, where higher scores indicate greater contribution to model predictions.}
    \label{fig:token_ranking}
\end{figure*}

Figure~\ref{fig:token_ranking} illustrates token ranking scores for five representative samples from the MMLU Mathematics domain evaluated with GPT-OSS-20B. The results show a highly non-uniform distribution of token importance across the input sequence, where tokens corresponding to mathematical expressions and key terms consistently receive markedly higher scores.

\begin{figure*}[t]
    \centering
    \includegraphics[width=\textwidth]{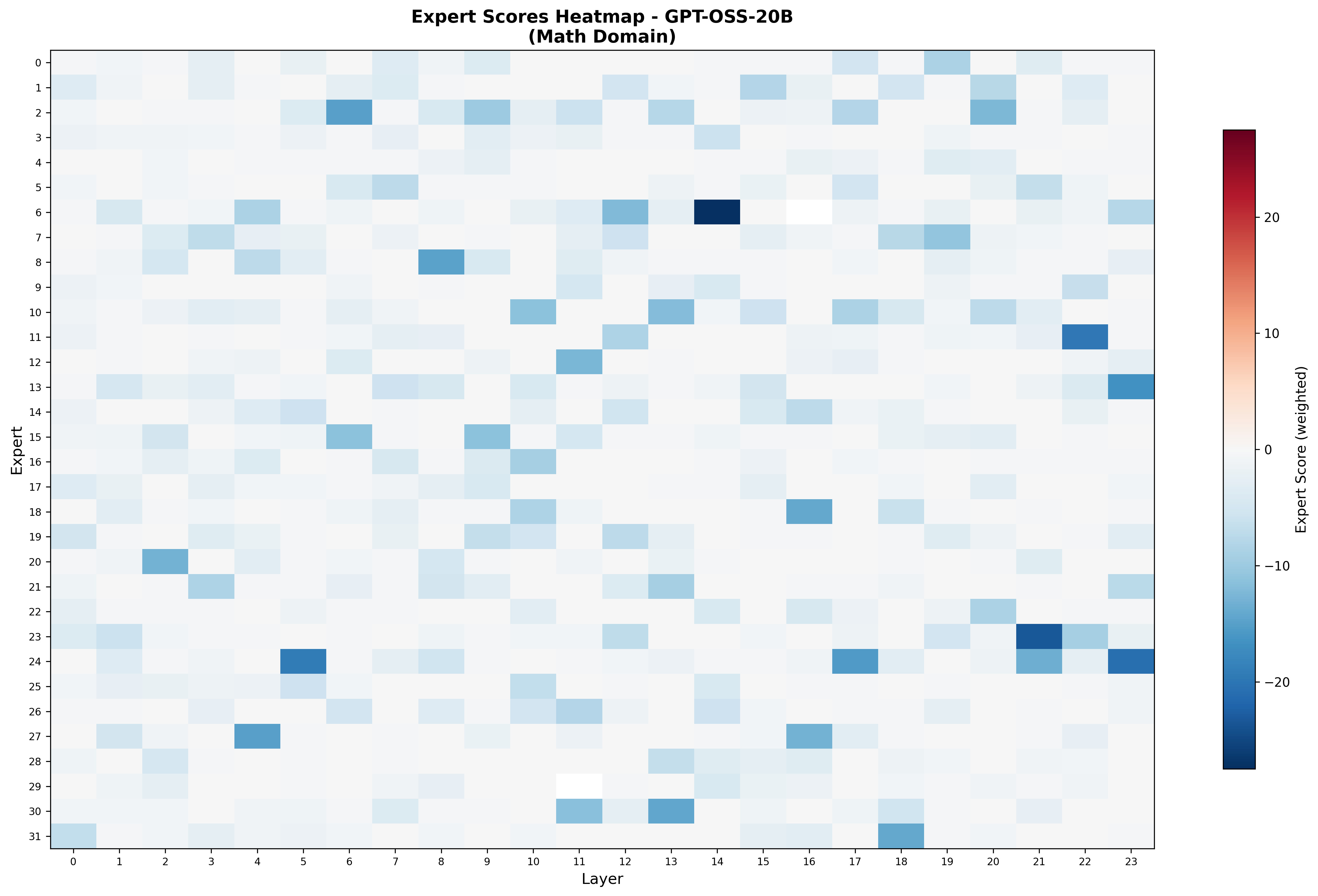}
    \caption{Domain-specific expert scores for \textbf{GPT-OSS-20B} on the Mathematics domain. Higher magnitudes indicate stronger domain specialization. Best viewed in color.}
    \label{fig:gpt20b}
\end{figure*}

Figure~\ref{fig:gpt20b} to Figure~\ref{fig:qwen_instruct} present heatmap visualizations of domain-specific expert scores for four MoE-based LLMs on the Mathematics domain. Across all models, we observe that domain-specific experts are not uniformly distributed; instead, certain layers exhibit clusters of highly specialized experts (indicated by higher magnitudes), while others contain more domain-agnostic experts. This non-uniform distribution provides empirical evidence supporting the existence of domain-specific experts and motivates our proposed steering approach.

\begin{figure*}[t]
    \centering
    \includegraphics[width=\textwidth]{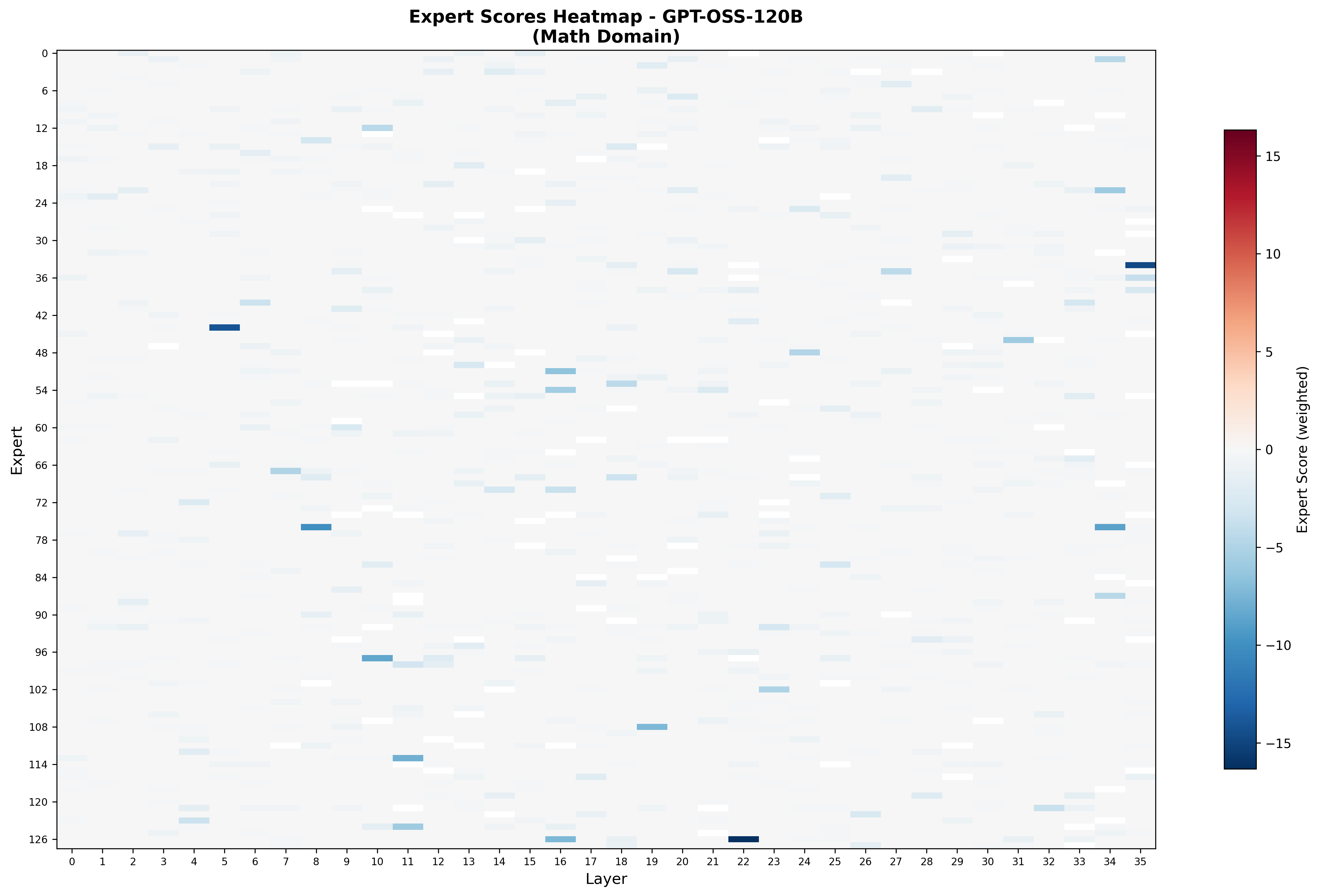}
    \caption{Domain-specific expert scores for \textbf{GPT-OSS-120B} on the Mathematics domain. Higher magnitudes indicate stronger domain specialization. Best viewed in color.}
    \label{fig:gpt120b}
\end{figure*}

\begin{figure*}[t]
    \centering
    \includegraphics[width=\textwidth]{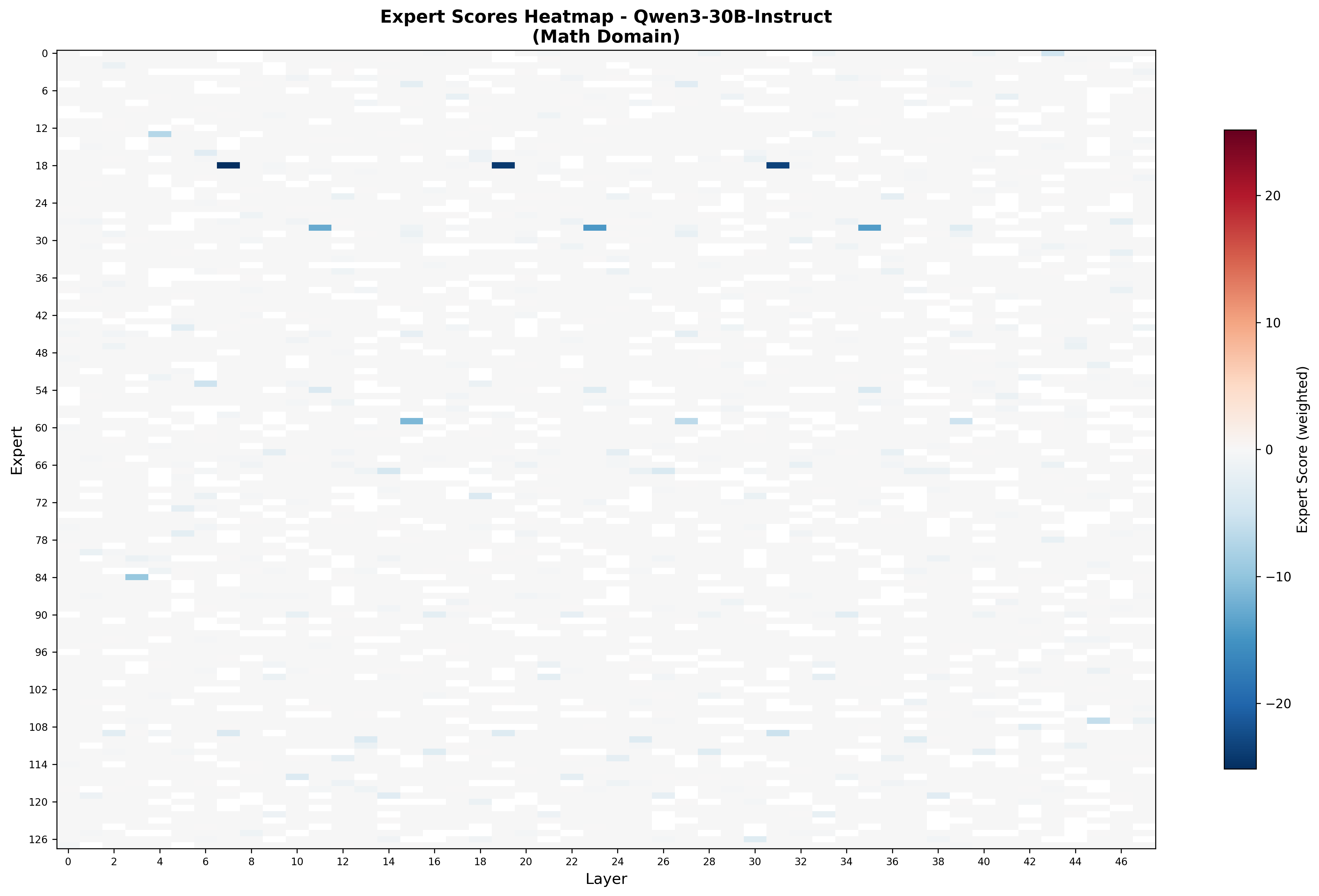}
    \caption{Domain-specific expert scores for \textbf{Qwen3-30B-Instruct} on the Mathematics domain. Higher magnitudes indicate stronger domain specialization. Best viewed in color.}
    \label{fig:qwen_instruct}
\end{figure*}

\begin{figure*}[t]
    \centering
    \includegraphics[width=\textwidth]{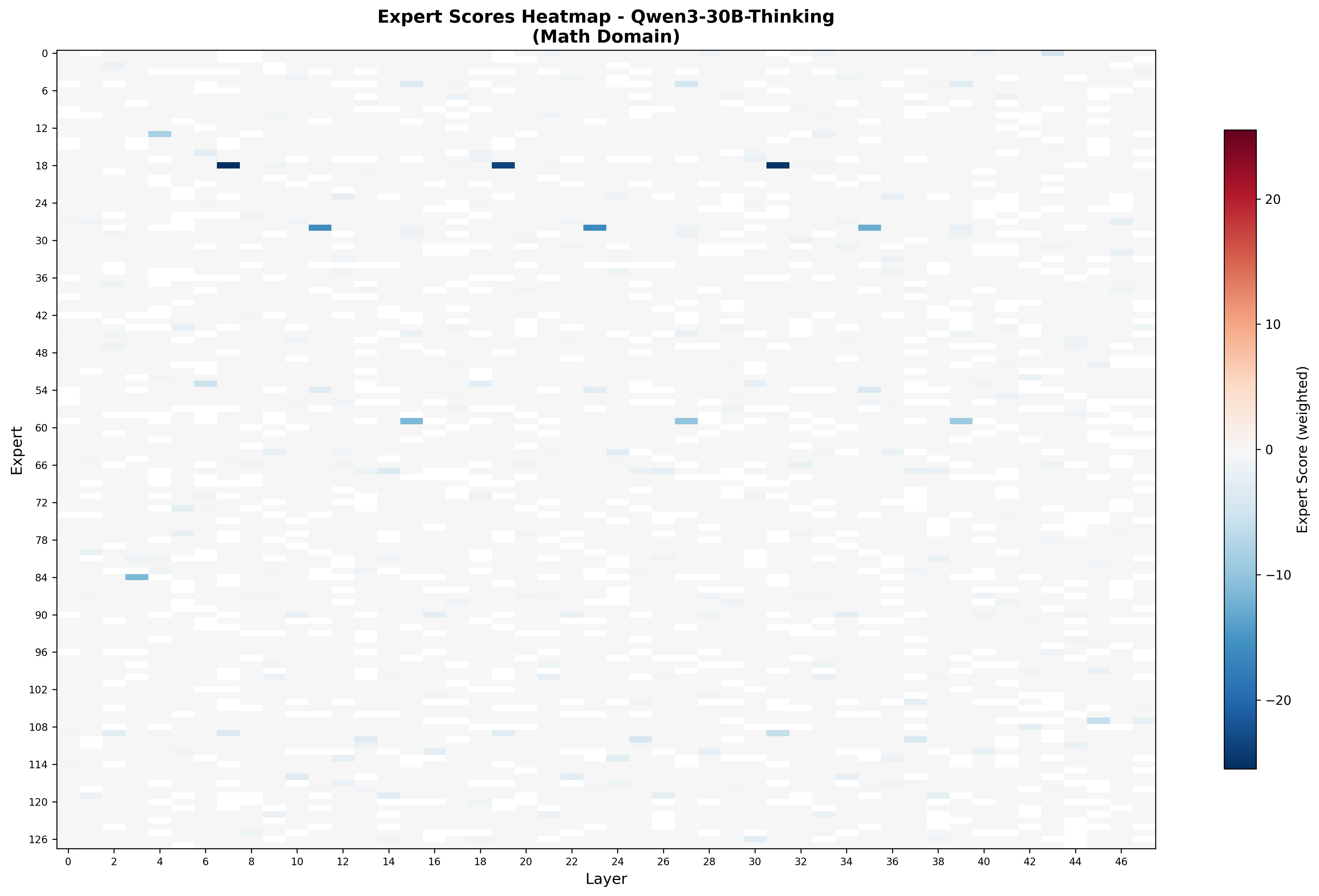}
    \caption{Domain-specific expert scores for \textbf{Qwen3-30B-Thinking} on the Mathematics domain. Higher magnitudes indicate stronger domain specialization. Best viewed in color.}
    \label{fig:qwen_instruct}
\end{figure*}

\subsection{Benchmarks Descriptions}
\label{app:add_imp}

Table~\ref{tab:model_specs} summarizes the architectural specifications of the ten MoE-based LLMs used in our experiments. The models span a wide range of scales, from 3.8B to 120B total parameters, with activated parameters ranging from 1.0B to 5.1B per token. The number of experts varies significantly across architectures, ranging from 16 (PhiMoE-Tiny) to 512 (Qwen3-Next), with Top-$K$ routing selecting between 2 and 10 experts per token.

\begin{table*}[t]
\centering

\resizebox{\textwidth}{!}{
\begin{tabular}{lcccccl}
\toprule
\textbf{Model} & \textbf{Params} & \textbf{Active} & \textbf{Experts} & \textbf{Top-$K$} & \textbf{Layers} & \textbf{HuggingFace Model ID} \\
 & \textbf{(B)} & \textbf{(B)} & \textbf{($N$)} & & & \\
\midrule
PhiMoE-Tiny & 3.8 & 1.1 & 16 & 2 & 32 & \texttt{microsoft/Phi-tiny-MoE-instruct} \\
OLMoE & 7.0 & 1.0 & 64 & 8 & 16 & \texttt{allenai/OLMoE-1B-7B-0924} \\
Qwen1.5-MoE & 14.3 & 2.7 & 60 & 4 & 24 & \texttt{Qwen/Qwen1.5-MoE-A2.7B} \\
DeepSeek-MoE & 16.0 & 2.8 & 64 & 6 & 28 & \texttt{deepseek-ai/deepseek-moe-16b-base} \\
GPT-OSS-20B & 20.0 & 3.6 & 32 & 4 & 24 & \texttt{openai/gpt-oss-20b} \\
ERNIE-4.5 & 21.0 & 3.0 & 64 & 6 & 28 & \texttt{baidu/ERNIE-4.5-21B-A3B-Thinking} \\
Qwen3MoE-Instruct & 30.0 & 3.0 & 128 & 8 & 48 & \texttt{Qwen/Qwen3-30B-A3B-Instruct-2507} \\
Qwen3MoE-Think & 30.0 & 3.0 & 128 & 8 & 48 & \texttt{Qwen/Qwen3-30B-A3B-Thinking-2507} \\
Qwen3-Next & 80.0 & 3.0 & 512 & 10 & 48 & \texttt{Qwen/Qwen3-Next-80B-A3B-Thinking} \\
GPT-OSS-120B & 120.0 & 5.1 & 128 & 4 & 36 & \texttt{openai/gpt-oss-120b} \\
\bottomrule
\end{tabular}%
}
\caption{Architectural specifications of the MoE models used in our experiments. We report the Total Parameters, Activated Parameters per token, Total Number of Experts, Experts selected per token (Top-$K$), Number of Layers, and the corresponding HuggingFace Model ID.}
\label{tab:model_specs}
\end{table*}

Table~\ref{tab:benchmark_meta} summarizes the evaluation benchmarks used in our experiments. We assess model performance across three challenging benchmarks: MMLU-Pro (12K multi-domain questions), GPQA Diamond (198 PhD-level science questions), and AIME (30 competition-level mathematics problems).

\begin{table*}[t]
\centering
\resizebox{\textwidth}{!}{
\begin{tabular}{llcl}
\toprule
\textbf{Benchmark} & \textbf{Domain} & \textbf{Size} & \textbf{Description} \\
\midrule
MMLU-Pro & STEM, Law, etc. & 12K & 10-choice questions; minimizes random guessing \\
GPQA Diamond & Biology, Physics, Chemistry & 198 & PhD-level difficulty; 65\% expert accuracy \\
AIME & Mathematics & 30 & Competition-level; integer answers (0-999) \\
\bottomrule
\end{tabular}%
}
\vspace{5pt}
\caption{Summary of meta data for the evaluation benchmarks. \textbf{Size} denotes the total number of questions in the dataset (or specific subset used).}
\label{tab:benchmark_meta}
\end{table*}





\subsection{Implementation Details}
\label{app:imp}

\noindent \textbf{Training-Free Setting.}\; For DSMoE steering experiments, we implement our method based on the publicly available RICE implementation~\cite{wang2025two}\footnote{\url{https://openreview.net/forum?id=x7fCiuCCAu}}. For \textit{GPT-OSS-120B}, experiments are conducted on two NVIDIA H200 GPUs, while for \textit{GPT-OSS-20B} and \textit{Qwen3-MoE}, we utilize two NVIDIA H100 GPUs. Following RICE, our implementation leverages the vLLM library~\citep{kwon2023efficient}, which supports parallel model loading and inference. We upgrade to vLLM version 0.11.0 to ensure compatibility with recent MoE-based LLMs. All models and datasets used in this work are publicly available on Hugging Face, ensuring full reproducibility of our results.

\noindent \textbf{SFT Baseline.}\; We implement the SFT baseline using the open-source PEFT library~\citep{peft} with the following configuration: 3 training epochs, LoRA rank of 16, LoRA alpha of 32, and target modules including \texttt{q\_proj}, \texttt{k\_proj}, \texttt{v\_proj}, \texttt{o\_proj}, \texttt{gate\_proj}, \texttt{up\_proj}, and \texttt{down\_proj}. This configuration results in approximately 2.7\% trainable parameters relative to the total model size.

\end{document}